%% file: emnlp2022.tex
\pdfoutput=1

\documentclass[11pt]{article}

\usepackage{EMNLP2022}

\usepackage{times}
\usepackage{latexsym}

\usepackage[T1]{fontenc}

\usepackage[utf8]{inputenc}

\usepackage{microtype}

\usepackage{inconsolata}
\usepackage{amssymb}
\usepackage{amsmath}
\usepackage{url}
\usepackage{multirow}
\usepackage{enumitem}
\usepackage{graphicx}
\usepackage{caption}
\usepackage{subcaption}

\newtheorem{definition}{Definition}

\newtheorem{theorem}{Theorem}

\newcommand{\STAB}[1]{\begin{tabular}{@{}c@{}}#1\end{tabular}}

\title{On the Effectiveness of Automated Metrics for Text Generation Systems}

\author{Pius von D\"{a}niken \and Jan Deriu 
\and Don Tuggener \and Mark Cieliebak \\
Centre for Artificial Intelligence\\
ZHAW School of Engineering \\
\texttt{\{vode,deri,tuge,ciel\}@zhaw.ch} \\
}
\begin{document}
\maketitle
\begin{abstract}

A major challenge in the field of Text Generation is evaluation, because we lack a sound theory that can be leveraged to extract guidelines for evaluation campaigns. In this work, we propose a first step towards such a theory that incorporates different sources of uncertainty, such as imperfect automated metrics and insufficiently sized test sets. The theory has practical applications, such as determining the number of samples needed to reliably distinguish the performance of a set of Text Generation systems in a given setting. 
We showcase the application of the theory on the WMT 21 and Spot-The-Bot evaluation data and outline how it can be leveraged to improve the evaluation protocol regarding the reliability, robustness, and significance of the evaluation outcome.
\end{abstract}

\section{Introduction}

The field of Text Generation is a subfield of Natural Language Processing~\cite{celikyilmaz2020evaluation}. We define text generation tasks as those where many different texts may constitute an optimal solution to a given problem. Examples are automated summarization, machine translation, dialogue systems, paraphrasing, caption generation, or natural language generation.

One unsolved issue in the field of Text Generation is the evaluation, be it human or automated evaluation. Human evaluation is more reliable but more cost and time intensive, and automated evaluation is erroneous but performed in a fraction of time and cost~\cite{amidei-etal-2019-use,hashimoto-etal-2019-unifying,celikyilmaz2020evaluation,deriu2021survey}. One of the main issues is the lack of theoretically founded guidelines when running an evaluation. For instance, how many samples are needed to be able to significantly distinguish the performance of two systems? Or how do we handle the errors made by automated metrics? Under which circumstances is it still possible to run an evaluation campaign that yields significant results? 
In this work, we make a first step towards developing such a theoretical foundation, which can be used as a guideline to answer the above questions. For this, we consider what we call \emph{binary metrics}. These are metrics that classify the output of a text generation system as being either adequate or inadequate. This allows us to measure the performance of a text generation system as the ratio of adequate responses it generates. Furthermore, it allows us to reason about the performance of the metric in terms of true positives and true negatives. 

\begin{figure}[t!]
    \centering
    \includegraphics[width=\columnwidth]{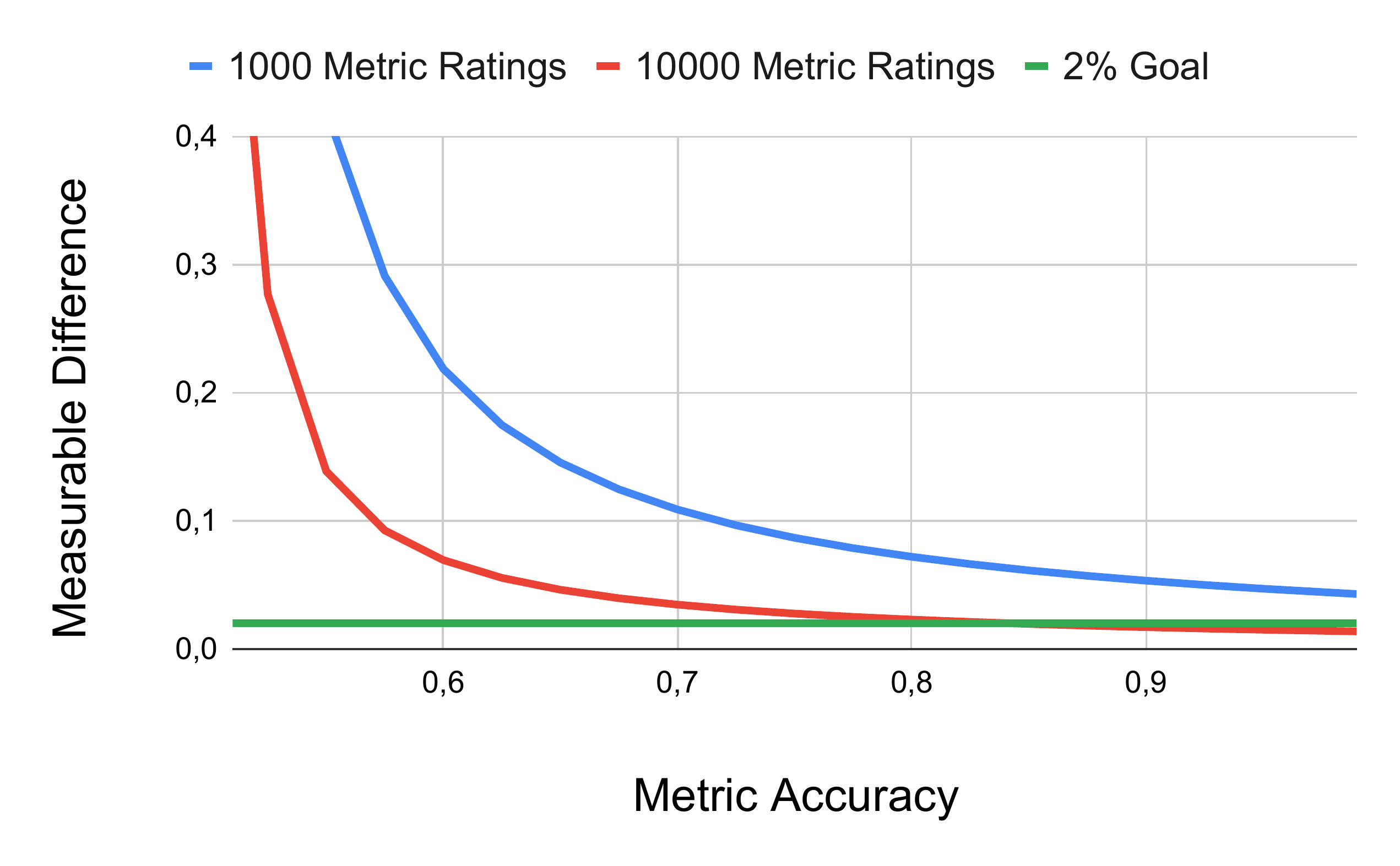}
    \caption{Measurable difference of the performance of two text generation systems depending on the accuracy of a binary metric. We add the $2\%$ line as discussed in the text.}
    \label{fig:intro}
\end{figure}

\begin{table}[ht!]
    \small
    \input{tables/alpha60_rho70_eta70_fixed}
    \caption{Mixed Case: Measurable difference for a metric with accuracy of $70\%$ depending on the number of human rating mixed with the number of automated ratings. The values discussed in the text are highlighted in bold.}
    \label{tbl:intro_a}
\end{table}

For this setting, we derive various theoretically founded guarantees and guidelines that can be used to run an evaluation campaign. For instance, consider Figure~\ref{fig:intro} (derived by our theory). If we assume a binary metric that has an accuracy of 70\%, and if we have access to 1000 automatically rated samples (blue line), then we can reliably distinguish between two text generation systems that have a difference in performance of 10 percentage points. To distinguish two systems with a smaller difference, for instance of 2\%, we would need a better metric and many more samples. That is, we need for instance a metric with an accuracy of at least 85\% and 10000 automatically rated samples by this metric.

Our theory provides analogous assessments of how many human evaluations are required to reliably distinguish text generation systems. When we say that the performance of two systems can be reliably distinguished, we mean that the difference in their performance is statistically significant. Similarly, a measurable difference in performance is one that leads to statistical significance given the experiment parameters.

In addition, our theory allows for the mix of human and automated evaluation. For this, consider Table~\ref{tbl:intro_a} where we depict the number of human and automatic ratings required by a metric with 70\% accuracy. For instance, to distinguish two text generators with 2 percentage points difference, we need either at least 5000 human ratings, or 2500 human ratings mixed with 10’000 automated ratings. 

Our theoretical framework allows us to design our evaluation with theoretical guarantees regarding the significance of the resulting measurements. Given a monetary budget and our theory, one can decide whether to invest in more human annotations, in developing better automated metrics, or in sampling more automated ratings. Our approach can also be used to showcase the limits of a given setting: for instance in Figure ~\ref{fig:intro}, we see that using only 1000 automated ratings leads to a minimally measurable difference of 4\% even with a perfect metric. 

In the remainder of the paper, we derive the theoretical framework for binary metrics and apply it to two showcases: the WMT-21 shared task~\cite{freitag-etal-2021-results} and the Spot-The-Bot evaluation~\cite{deriu-etal-2020-spot}. We analyse how well these evaluations adhere to the constraints imposed by our theory and demonstrate how the quality of the evaluations can be improved. To serve the community, we will release the formulas as code and as a web interface~\footnote{\url{https://github.com/vodezhaw/binary\_metric\_tool}} that allows practitioners to enter their evaluation settings and receive an analysis of the measurable differences in their settings.

\section{Definitions}
In this section, we introduce the basic definitions that we need for the derivations. First, we define the general setting of Text Generation, then we cover binary metrics, and finally we describe text generation systems. 

\subsection{General Setting}
\begin{definition}[Text Generation Environment]
\label{def:tge}
A \emph{text generation environment} is composed of a triple $\left \langle \mathcal{I}, \mathcal{O}, \Phi \right \rangle$, where $ \mathcal{I}$ denotes the set of inputs, $\mathcal{O}$ the output space, and $\Phi : \mathcal{I} \times \mathcal{O} \rightarrow  \left \{ 0, 1 \right \}$ an oracle that assess whether an output is adequate for a given input.  
\end{definition}

For instance, for Machine Translation $\mathcal{I}$ denotes all sentences in the source language and $\mathcal{O}$ all sentences in the target language, while for a chatbot $\mathcal{I}$ contains all dialogue contexts and $\mathcal{O}$ all possible responses in a dialogue. Note that $\mathcal{I}$ and $\mathcal{O}$ can be of infinite size. We regard $\Phi$ as an oracle that segments the output space for a given input into adequate and inadequate outputs~\footnote{In most real-world setting $\Phi$ is approximated with human ratings.}. 

\begin{definition}[Adequate Responses]
$\forall i \in \mathcal{I}$, we call $\mathcal{R}_+^i = \left \{ o \in \mathcal{O} | \Phi(i,o) = 1 \right \}$ the set of \emph{adequate responses} for input i, and $\mathcal{R}_{-}^i = \left \{ o \in \mathcal{O} | \Phi(i,o) = 0 \right \}$ the set of \emph{inadequate responses}.
\end{definition}

\subsection{Binary Metric}
In this work, we set our focus to binary metrics, i.e., metrics that classify the output of a text generation system as being either adequate or inadequate. The choice of binary metrics allows us to reason about the performance of a text generation (TG) system as the ratio of adequate responses\footnote{This lies in contrast with metrics that simply return a scalar value (e.g, BLEU~\cite{papineni-etal-2002-bleu}, COMET~\cite{rei-etal-2020-comet}, USR~\cite{mehri-eskenazi-2020-usr}) that is difficult to interpret. For instance, if BLEU returns a value of 0.34 for one system and 0.32 for the second system, can we really state that the first system is better than the second \cite{callison-burch-etal-2006-evaluating}? We can use these types of metrics to create binary metrics by selecting a threshold that defines the border between adequate and inadequate responses (e.g., all COMET values above 0.78 are regarded as adequate). This introduces errors, which can be measured.}.

We first define the notion of a binary metric, then we show what it means for a binary metric to be error-free or error-prone with regards to $\Phi$.

\begin{definition}[Binary Metric]
\label{def:binary_metric}
A \emph{binary metric $M_b$} is a function $M_b: \mathcal{I} \times \mathcal{O} \rightarrow \{0, 1\}$ which takes a pair of input and output, and returns either 0 or 1. We interpret the return of 1 as claiming that the output is an adequate output for the given input, and 0 claiming that the output is not adequate.
\end{definition}

Next, we define the notion of an error-free metric. That is, how we expect the metric to behave in the optimal case (i.e.\ its ability to replicate the oracle $\Phi$). 

\begin{definition}[Error-Free Binary Metric]
\label{def:consistent_binary}
$M_b^{*}$ is an \emph{error-free} binary metric $\iff \forall i \in \mathcal{I}, o \in \mathcal{O}: (M_b^{*}(i, o) = 1 \iff o \in\mathcal{R}_+^i)$.
\end{definition}

That is, an error-free binary metric always rates an adequate output as $1$ and an inadequate output as $0$. Since most metrics do not perform perfectly regarding $\Phi$, we formulate the cases where a metric makes mistakes and the calculation of its performance as follows.

\begin{definition}[$(\rho, \eta)$-optimal binary metric]
Let $\rho, \eta \in  [0, 1]$ and $M_b$ a binary metric. Then $M_b$ is a \emph{$(\rho, \eta)$-optimal binary metric} if 
$Pr[M_b(i, o) = 1 | o \in \mathcal{R}_+^i] = \rho$ and $Pr[M_b(i, o) = 0 | o \not\in \mathcal{R}_+^i] = \eta$. 
\end{definition}
That is, we define the performance of a binary metric as its probability to correctly classify an output as being adequate or not. Thus, the error of a binary metric can be assessed similar to the error of a binary classifier, i.e., $\rho$ is equivalent to the true positive ratio and $\eta$ to the true negative ratio. Note that $\rho=\eta=1$ defines an error-free binary metric, whereas all other cases are error-prone. In the case where $\rho$ and $\eta$ have the same value, $\rho = \eta$, this value is the accuracy of $M_{b}^{\rho, \eta}$. Note that in practise, $\rho$ and $\eta$ must be estimated from data.

\subsection{Text Generation}
We define a text generation system as a function that takes an input from the input-space and generates an output. 

\begin{definition}[(Optimal) Text Generator]
\label{def:rg_DEF}
A \emph{Text-Generator (TG)} is a mapping $\pi : \mathcal{I} \to \mathcal{O}$ which generates for each input $i$ an output $o$. A TG is \emph{optimal} $\iff \forall i \in \mathcal{I}: \pi(i) \in R_+^i$
\end{definition}

Next, we introduce the notion of an imperfect text-generator. There are many different ways the errors of a TG can be modeled. We model it as its capability of generating adequate responses. 

\begin{definition}[$\alpha$-optimal TG]\label{def:alpha}
Let $\pi$ be a TG and $\alpha \in  [0, 1]$. Then $\pi$ is an  \emph{$\alpha$-optimal} TG if $Pr[\pi(i) \in \mathcal{R}_+^i] = \alpha$ for all $i \in \mathcal{I}$.
\end{definition}
That is, the probability of the text generation system to generate an adequate output is denoted as $\alpha$. The task of a binary metric is to estimate the $\alpha$ value of a TG system, which has a concrete meaning: Assume that we compare two systems, where $\alpha^{\pi_1} = 0.5$, and $\alpha^{\pi_2} = 0.49$, then these numbers have a clear semantic: $\pi_1$ outputs an adequate output in 50\% of cases and $\pi_2$ in 49\% of cases. Thus, one system generates adequate outputs more often than the other. We denote the difference in performance as $\epsilon$.
In the following, we will use $\alpha^{\pi}$ to denote the rate at which a system $\pi$ generates adequate responses, and $\pi^{\alpha}$ to refer to a system which is $\alpha$-optimal.



\section{Theory: Estimating $\alpha$ with Binary Metrics}\label{sec:theory}
In this section, we show how binary metrics can be used to estimate the performance $\alpha$ of text generation systems. For the remainder of the text, assume that $\mathcal{T}_{\Phi} = \{(i_j, o_j, r_j^*) | 1 \leq j \leq n_{\phi}\}$ is a set of input-output rating triples of size $n_{\phi}$, where $i_j$ are inputs, $o_j = \pi^{\alpha}(i_j)$ denotes the output generated by an $\alpha$-optimal TG system for input $i_j$, and $r_j^* = M_b^*(i_j, o_j)$ denotes the error-free rating of the $j^{th}$ input-output-pair. Analogously, let $\mathcal{T}_M = \{(i_j, o_j, r_j) | 1 \leq j \leq n_M\}$ be a set of input output rating triples of size $n_M$, where $r_j = M_b^{\rho,\eta}(i_j, o_j)$ denotes the rating of an error-prone $(\rho,\eta)$-optimal binary metric.

We consider three different cases: 1) the error-free case, 2) the error-prone metric case, and 3) the mixed case. The error-free case is where we have access to $r_j^*$. For instance, we can interpret human evaluation as an example of the error-free case. In the error-prone metric case, we have access only to an  $(\rho,\eta)$-optimal binary metric. Finally, the mixed case is a novel approach that leverages error-free ratings, which are usually costly to obtain, with error-prone ratings, which are cheaper but are needed en-masse for automated metrics with low $\rho$ and $\eta$ values, as we will see. Usually, in evaluation campaigns, either the first or second setting is applied. 

We apply a Bayesian approach to estimate $\alpha$ by treating it as a random variable, which allows us to model various sources of uncertainty stemming from $\alpha$, $\rho$ and $\eta$, which all need to be estimated from data. The full derivations are given in Appendix \ref{app:derivation}.

\subsection{Error-Free Case}
\label{sec:consistent_case}
Here, we start with the most simple case and introduce the formula to estimate $\alpha$ given error-free ratings $r_j^*$. Given $n_{\phi}$ error-free ratings, $\alpha$ is estimated by $\Tilde{\alpha} = \frac{n_+}{n_{\phi}}$, where $n_+ = \sum_{i=1}^{n_{\phi}} r_j^*$. This formula can be derived via the frequentist approach or the Bayesian. For the Bayesian approach, we assume a uniform prior over $\alpha$ (i.e. $\alpha \sim Beta(1,1)$). The resulting posterior distribution for $\alpha$ given $n_+$ is:
\begin{equation}
    \begin{aligned}
        &P(\alpha| N_+ = n_+) \propto P(N_+ = n_+|\alpha)*P(\alpha) \\
        &\propto Beta(n_+ + 1, n_{\phi} - n_+ + 1)
    \end{aligned}
\end{equation}
and the value of $\alpha$ is estimated using the mode of $Beta(n_+ + 1, n_{\phi} - n_+ + 1)$, which corresponds to $\frac{n_+}{n_{\phi}}$.


\subsection{Error-Prone Metric Case}
In the error-prone metric case, the probability that $r_j = 1$ depends on $\rho$ and $\eta$. Hence, if $r_j = 1$, we cannot assume that $r_j^*=1$ as well, since the binary metric can be error-prone. For the error-prone setting, we consider two cases, one where $\rho$ and $\eta$ are provided (e.g. from an earlier evaluation campaign), and one where $\rho$ and $\eta$ must be estimated from data (i.e., from comparison to error-free ratings). 

\subsubsection{Provided $\rho,\eta$}
Here, we assume that the exact values of $\rho$ and $\eta$ are known. The probability that the binary metric returns a positive label is thus given by:
\begin{equation}
 \begin{aligned}
   P(r_j = 1) = \alpha\rho + (1-\alpha)(1-\eta)
   \end{aligned}
\end{equation}

From this, we derive the formula to estimate $\alpha$ using the Bayesian formulation.
\begin{theorem}[Estimate $\alpha$ with error-prone metric]
\label{th:measure_alpha_inc}
Let $m_+ = \sum_{i=1}^{n_M} r_j \sim Binom(P(r_j = 1), n_M)$ be the number of pairs $i_j, o_j$ rated as adequate $M_b^{\rho,\eta}(i_j, o_j) = 1$. Then we estimate $\alpha$ by computing the mode of the following distribution:
\begin{equation}
\small
 \begin{aligned}
   &P(\alpha | M_+ = m_+, \rho, \eta)  \\
   &\propto P(M_+ = m_+|\alpha , \rho, \eta)P(\alpha)
   \end{aligned}
\end{equation} 
If we assume a uniform prior of $\alpha$, i.e., $P(\alpha) \sim U(0,1)$, this reduces to:
$\Tilde{\alpha} = \frac{\frac{m_{+}}{n_M} + \eta - 1}{\rho + \eta - 1}$
\end{theorem}

Note that the above formulation does not allow for $\rho + \eta = 1$, in which case our estimator would be undefined. In the following we will assume that $\rho + \eta > 1$. This is a relatively safe assumption since in the case where $\rho + \eta < 1$, we can derive a new metric $M_{b}^{\rho', \eta'}$ by flipping the predictions of $M_{b}^{\rho, \eta}$:
$M_{b}^{\rho', \eta'}(i, o) = 1 - M_{b}^{\rho, \eta}(i, o)$. In this case $\rho' + \eta' = (1 - \rho) + (1 - \eta) = 2 - (\rho + \eta) > 1$.

\subsubsection{Estimated $\rho,\eta$}
\label{sec:est_rho_eta}
Here, we assume that $\rho$ and $\eta$ must be estimated from data, which introduces uncertainty. In our case, we estimate $\rho$ and $\eta$ from error-free ratings (i.e., how well the error-prone metric agrees with the error-free ratings). In practise, the error-free assessments stem from human annotations, which are regarded as the ground truth. To weave the estimation of $\rho$ and $\eta$ into the Bayesian framework, we treat them as random variables. For this, assume that we have access to a dataset $T_{\rho,\eta} = \{(i_j, o_j, r_j^*, r_j) | 1 \leq j \leq M\}$ of both error-free and error-prone ratings for pairs of inputs and outputs. Denote $\mathcal{T}_{\rho,\eta}^+ = \{(i_j, o_j) | r_j^* = 1\}$ as the set of true positive samples, and $\mathcal{T}_{\rho,\eta}^- = \{(i_j, o_j) | r_j^* = 0\}$ as the set of true negative samples. 
Thus, assuming a uniform prior over $\rho$, we apply the same reasoning as in Section~\ref{sec:consistent_case} to compute the posterior distribution  $\rho \sim Beta(m^{TP} + 1, |\mathcal{T}_{\rho,\eta}^+| - m^{TP} + 1)$, where $m^{TP}$ denotes the number of true positive samples, rated as positive by $M_b^{\rho,\eta}$. Analogously, $\eta \sim Beta(m^{TN} + 1, |\mathcal{T}_{\rho,\eta}^-| - m^{TN} + 1)$, where $m^{TN}$ denotes the number of true negative samples, rated as negative by $M_b^{\rho,\eta}$. Note that to estimate $\rho$ and $\eta$, having a large sample size for both $\mathcal{T}_{\rho,\eta}^+$ and $\mathcal{T}_{\rho,\eta}^-$ is important, otherwise the estimation of $\rho$ or $\eta$ would have a higher uncertainty. 

To incorporate the uncertainty of $\rho$ and $\eta$ into the estimation of $\alpha$, we need to marginalize $\rho$ and $\eta$ from the joint likelihood $P(m_+,\rho,\eta|\alpha)$ to get $P(m_+|\alpha)$. 
\begin{theorem}[Est. $\alpha, \rho, \eta$ with error-prone metric]
Let $m_+ = \sum_{i=1}^n r_j \sim Binom(P(r_j = 1), n)$ be the number of samples rated positively by $M_b^{\rho,\eta}$. Then we estimate $\alpha$ by computing the mode of the following distribution:
\begin{equation}
\small
 \begin{aligned}
   P(\alpha |M_+ = m_+) \propto P(M_+ = m_+|\alpha)P(\alpha) \\
   \propto P(\alpha)\int_{0}^{1}\int_{0}^{1} p(m_{+} | \alpha, \rho, \eta)p(\rho)p(\eta)\mathrm{d}\rho\mathrm{d}\eta
   \end{aligned}
\end{equation}
\end{theorem}

Note that we are not aware of a closed form solution for the above distribution and the computation of the mode. Thus, we approximate the solution using numerical methods in practise (See Appendix~\ref{app:sim}). 

\subsection{Mixed Case}
\label{sec:mixed_case}
The mixed case combines the error-free and the error-prone cases. Here, we assume that we are given a small number of error-free samples (human annotations), which are costly to obtain, and a larger set of error-prone samples (ratings by an automated metric), which are easier to obtain~\footnote{Note that our setting also allows for $\mathcal{T}_{\Phi} \subseteq \mathcal{T}_M$.}.

\begin{theorem}[Mixed $\alpha$ estimation]
Let $n_+ = \sum_{i=1}^{|\mathcal{T}_{\Phi}|} r_j^* \sim Binom(\alpha, |\mathcal{T}_{\Phi}|)$ the number of samples where $M_b^* = 1$, and $m_+ = \sum_{i=1}^n r_j \sim Binom(P(r_j = 1), |\mathcal{T}_M|)$ be the number of samples where $M_b^{\rho,\eta} = 1$.  Then we estimate $\alpha$ by computing the mode of the following distribution:
\begin{equation}\label{eq:fusion}
\small
 \begin{aligned}
   &P(\alpha |M_+ = m_+, N_+ = n_+) \\
   &\propto P(M_+ = m_+, N_+ = n_+|\alpha)P(\alpha) \\
   &\propto P(\alpha|N_+ = n_+) \\
   &\times\int_{0}^{1}\int_{0}^{1} P(M_+ = m_{+} | \alpha, \rho, \eta)p(\rho)p(\eta)\mathrm{d}\rho\mathrm{d}\eta
   \end{aligned}
\end{equation}
\end{theorem}

Note that the difference to the error-prone case is that $P(\alpha)$ is replaced by $P(\alpha|n_+)$, which can be expressed by a closed form beta distribution (see Section \ref{sec:consistent_case}). Thus, we can compute the mixed case by first computing the error-free case to get an initial estimate of $\alpha$, and then estimate the error-prone case. More generally, this approach lets us also combine ratings from multiple different error-prone metrics by applying Equation \ref{eq:fusion} iteratively. One would plug in the posterior from one metric as the prior for the next. 

Having outlined the estimation of $\alpha$ for different scenarios, we now show how they can be used to determine the minimal number of samples needed to distinguish TGs in a significant manner.

\section{Minimal Number of Samples Needed to Make Reliable Distinctions between TG Systems}
 
We now come back to the main question of this paper: how many samples are needed to be able to significantly distinguish the performance of two text generation systems? The intuition is that the closer the performance of the two TG  systems is, the more samples are needed. Thus, we investigate the setting where their difference in performance $|\alpha^{\pi_1} - \alpha^{\pi_2}|= \epsilon$ is small. Using the formulas from Section~\ref{sec:theory}, we can compute the estimates shown in Table~\ref{tbl:intro_a}. There are seven variables involved in this computation:

\begin{itemize}[noitemsep,topsep=0pt]
    \item $\rho$ and $\eta$ denote the (unknown) performance of the automated binary metric. The better it is, the less samples are needed.
    \item $\alpha$ denotes the (unknown) performance of the TG system to be evaluated.
    \item $\gamma$ as the significance level that is wished to be achieved.
    \item $|\mathcal{T}_{\Phi}|$ denotes the size of the set of rated input-output pairs that stem from a error-free binary metric.
    \item $|\mathcal{T_M}|$ denotes the size of the set of rated input-output pairs that stem from an error-prone binary metric. 
    \item $|\mathcal{T_{\rho, \eta}}|$ denotes the set of samples needed to estimate $\rho$ and $\eta$.
\end{itemize}

To compute if one system is significantly better, the probability of one system being better than the other must be compared to the significance level (e.g., 0.05).
We compute the probability that $\alpha_1 > \alpha_2$ as follows:
\begin{equation}\label{eq:bigger_dist}
\small
    \begin{aligned}
        P(\alpha_1 > \alpha_2) = \int_{0}^{1}\int_{\alpha_2}^{1}p(\alpha_1)p(\alpha_2)\mathrm{d}\alpha_1\mathrm{d}\alpha_2
    \end{aligned}
\end{equation}
The difference between $\pi^{\alpha_1}$ and $\pi^{\alpha_2}$ is significant at the $\gamma$-level if $P(\alpha_1 > \alpha_2) < 1 - \frac{\gamma}{2}$ or
$P(\alpha_1 > \alpha_2) < \frac{\gamma}{2}$.

Equation \ref{eq:bigger_dist} holds for any two random variables. In the particular case of normal distributions this is a reformulation of a two-sided z-test of the null hypothesis that both variables have the same mean.
Equation \ref{eq:bigger_dist} is therefore applicable to all the three cases of $\alpha$ estimation (i.e., error-free, error-prone, and mixed) by inserting the posterior distributions.

By applying normal approximations for $p(\alpha_1)$ and $p(\alpha_2)$, and using simulations we can compute the minimal distinguishable difference $\epsilon$ for a given set of fixed parameters. The details of the simulations are given in Appendix~\ref{app:sim}. 


\section{Showcases: Application in Practise} 
In order to show that the theoretical findings translate to practical applications, we apply our theory to two real-world settings: the WMT21 metric shared task~\cite{freitag-etal-2021-results} and the Spot-The-Bot data~\cite{deriu-etal-2020-spot}. Since the two tasks have significantly different settings (e.g., machine translation and dialogue systems, different types of human annotations, and different types of metrics) this shows that our theory is applicable to a variety of text generation tasks. The showcases highlight the different dimensions that can be manipulated when designing an evaluation. In showcase 1, we highlight the number of ratings needed, whereas, in showcase 2, we focus on the influence of the metric performance.

\subsection{Showcase 1: WMT Metrics Shared Task}\label{sec:wmt_sc}
For the WMT21 Metrics shared task, the authors evaluated the performance of 15 automated metrics by comparing their ratings to human ones on the output of several MT systems and several language pairs. In this work, we only focus on the English to German language pair and the news domain, where seven machine translation systems were evaluated. The data provided by the shared task can be expressed as follows using our notation: We regard the expert human multidimensional quality metrics (MQM)~\cite{lommel2014multidimensional} annotations as our error-free ratings. We binarize the scalar output of this metric by stating that only translations without any mistakes are regarded as adequate (i.e., $o \in \mathcal{R}_+^i \iff MQM(i, o) = 0$). This means only responses that have been judged as being completely correct by all annotators are considered adequate. For this setting there are $|\mathcal{T}_{\Phi}| = 527$ error-free annotated samples for each machine translation system. We can reuse these annotations to estimate $\rho$ and $\eta$, thus,  $|\mathcal{T}_{\rho,\eta}| = 527$~\footnote{Note that we estimate $\rho$ and $\eta$ for each machine translation system separately since we noted that most trained metrics have different performances depending on the various machine translation systems. See Appendix \ref{app:roc}.}. For the error-prone metric outputs, WMT provides $|\mathcal{T}_M| = 1000$ samples for each machine translation system and each error-prone metric. For the error-prone metrics, we use \emph{BleuRT}~\cite{sellam-etal-2020-bleurt} as the metric with the highest $\rho$ and $\eta$ estimates, and \emph{SacreBLEU}~\cite{post-2018-call} as the most popular metric. We consider three machine translation systems: FacebookAI (FBAI)~\cite{tran-etal-2021-facebook}, VolcTrans-GLAT (VT)~\cite{qian-etal-2021-volctrans}, and HuaweiTSC (HU)~\cite{wei-etal-2021-hw}, which have the most interesting combinations of performance (the full Tables are in Appendix~\ref{app:sc_tbl}).

\subsubsection{WMT: Theoretical Bounds of $\epsilon$}

\begin{figure}[t!]
    \centering
    \includegraphics[width=\columnwidth]{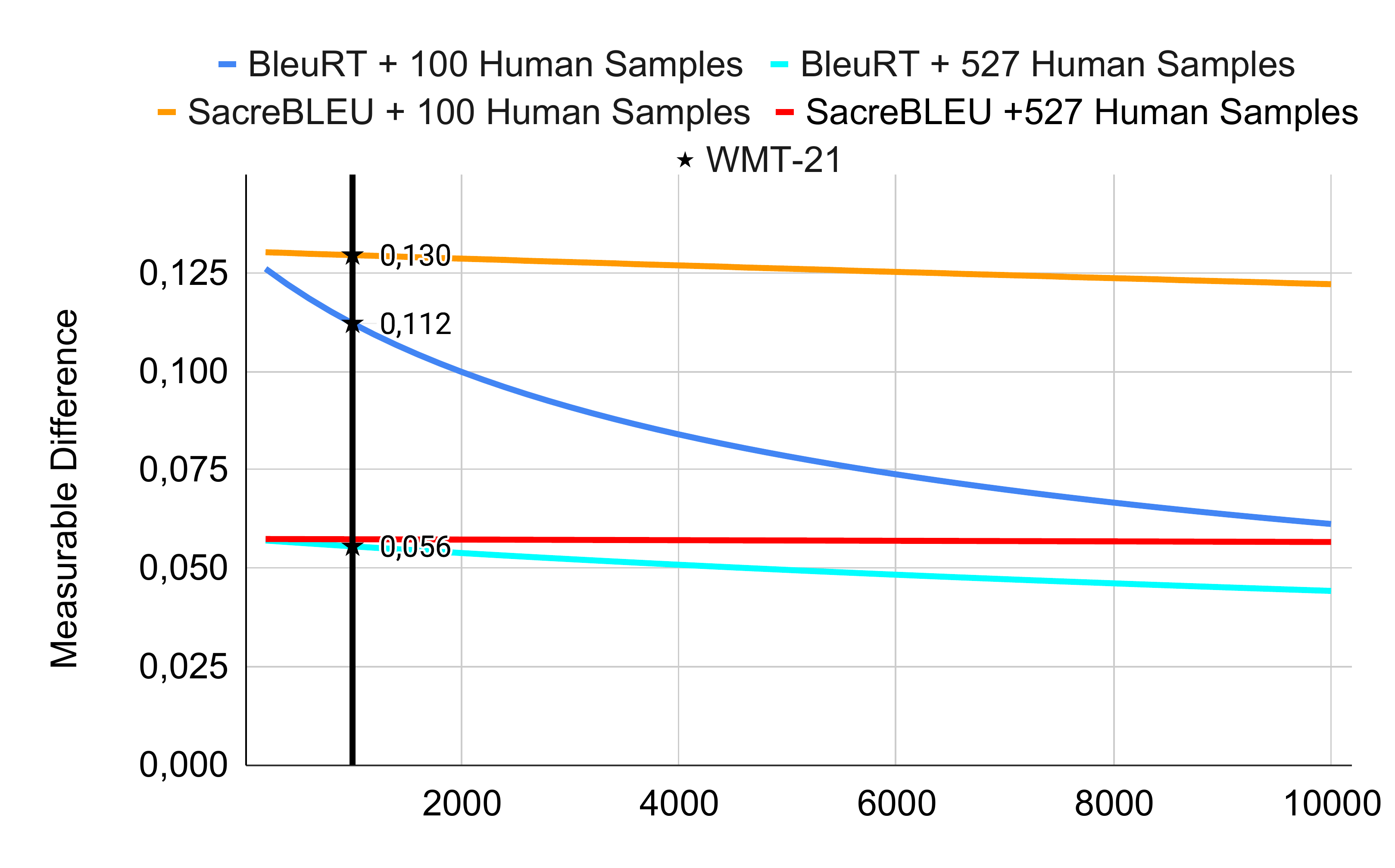}
    \caption{The measurable $\epsilon$ depending on $|\mathcal{T}_M|$ (x-Axis) for the BleuRT scenario ($\rho=\eta=0.6$) and the SacreBLEU scenario ($\rho=\eta=0.52$) and for $|\mathcal{T}_{\Phi}| = 100~\text{or}~527$. The vertical line shows the WMT setting with $|\mathcal{T}_M|=1000$.}
    \label{fig:wmt_theo} 
\end{figure}

Here, we showcase the theoretical bounds of the $\epsilon$ values that can be distinguished significantly depending on the number of ratings and the performance of the metrics. We consider \emph{BleuRT} with an estimated $\rho = \eta \approx 0.6$ (see Section~\ref{sec:est_rho_eta} on how to compute these estimates), \emph{SacreBLEU} with $\rho=\eta \approx 0.52$ and the performances of the machine translation systems are around $\alpha \approx 0.65$ (see section~\ref{sec:mixed_case} on how to compute the estimate). Figure~\ref{fig:wmt_theo} shows the theoretical $\epsilon$ values that can be distinguished for various values of $|\mathcal{T}_M|$ and $|T_{\Phi}|$. For instance, with 527 error-free ($|\mathcal{T}_{\Phi}|$) and 1000 error-prone samples ($|\mathcal{T}_M|$), we can distinguish an $\epsilon$ of $5.6\%$ for both \emph{BleuRT} and \emph{SacreBLEU}. Thus, the impact of the automated metrics is low for higher number of human ratings. However, for $\mathcal{T}_0 = 100$ the impact of the metric performance is larger: $\epsilon = 0.112$ vs. $\epsilon=0.13$. The effect is even larger with access to more automated ratings. Thus, using 10000 \emph{BleuRT} ratings with 100 human ratings allows to distinguish the same $\epsilon$ as with 527 human ratings and 1000 \emph{SacreBLEU} ratings, which is much costlier. 

\subsubsection{WMT: Practical Results}
\begin{table}[t!]
\centering
\small
\input{tables/showcase1/bleurt_full_test}
\caption{Predicted WMT21 evaluation using BleuRT and SacreBLEU on three machine translation systems.}
\label{tbl:wmt_prac_full} 
\end{table}

Here we analyse the results obtained when applying the theoretical framework to real data to estimate $\alpha$, and assess whether the pairwise differences are significant or not. 
Table~\ref{tbl:wmt_prac_full} shows the results for four scenarios: using all 527 error-free ratings, using only 100 error-free ratings (low-cost scenario), using 100 error-free ratings with an additional 1000 error-prone ratings from SacreBLEU, and using 100 error-free ratings with an additional 1000 error-prone ratings from BleuRT. The results include for each pair of systems the estimated $\epsilon$ values, and the probability that the first TG system is better than the second system. In the first scenario, we see that FBAI and VT cannot be significantly distinguished, which is consistent with the theory that states only $\epsilon > 0.057$ can be distinguished (see Figure~\ref{fig:wmt_theo}), whereas the other system pairs can be distinguished. In the second scenario, we reduce the number of error-free samples to only $|\mathcal{T}_{\Phi}| = 100$, which makes all the TG systems non distinguishable from each other. Again, this is consistent with the theory that states only $\epsilon > 0.131$ can be distinguished using 100 consistent samples. When we add error-prone ratings, the probabilities of the first TG being better than the second increase, however not enough to be significantly distinguishable. This goes for both automated metrics, which is still consistent with the theory. The problem lies in the fact that the performance of the automated metrics is too low to have a strong impact on the evaluation. For instance, the theory predicts that using $10'000$ error-prone SacreBLEU samples will only lead to being able to distinguish $\epsilon > 0.120$. In this setting, adding even more error-prone samples will not help (even with $|\mathcal{T}_M| = 10^9$), since the uncertainty of $\rho$ and $\eta$ is too high due to $|\mathcal{T}_{\rho,\eta}| = 527$.

Thus, the practical application shows that the outcomes using real data is consistent with the theory. Unfortunately, the setting does not allow to distinguish FBAI and VT. For this more error-free ratings are needed, or better metrics. 

\subsection{Showcase 2: Spot The Bot (STB)} \label{sec:stb_sc}
For the second show case, we use the Spot The Bot (STB) data, where dialogues between two dialogue systems are sampled and humans classified each interlocutor to be a human or a bot. STB contains pairwise ratings for six dialogue systems. In our setting, we use three of them: Blenderbot (BL)~\cite{roller-etal-2021-recipes}, Lost in Conversation~\footnote{\url{https://github.com/atselousov/transformer_chatbot}} (LiC), and KVMemNN (KV)~\cite{dinan2020second}.  In this setting the error-free metric is the (aggregated) human judgment, which is already binary. We consider a response as adequate if all annotators labelled it as coming from a human. For the error-prone metric, we use the USR~\cite{mehri-eskenazi-2020-usr} metric, which is also a scalar metric that we binarize with a threshold\footnote{See Appendix \ref{app:roc}.}. The STB dataset yields $|\mathcal{T}_{\Phi}| = |\mathcal{T}_{\rho,\eta}| \approx 600$ error-free ratings per dialogue system. For creating $\mathcal{T}_M$, we sample new pairwise dialogues and let USR rate each turn of the dialogue. This yields $|\mathcal{T}_M| = 10'000$ samples per dialogue system. 

\begin{figure*}[!t]
     \centering
     \begin{subfigure}[b]{0.3\textwidth}
         \centering
         \includegraphics[width=1.2\textwidth]{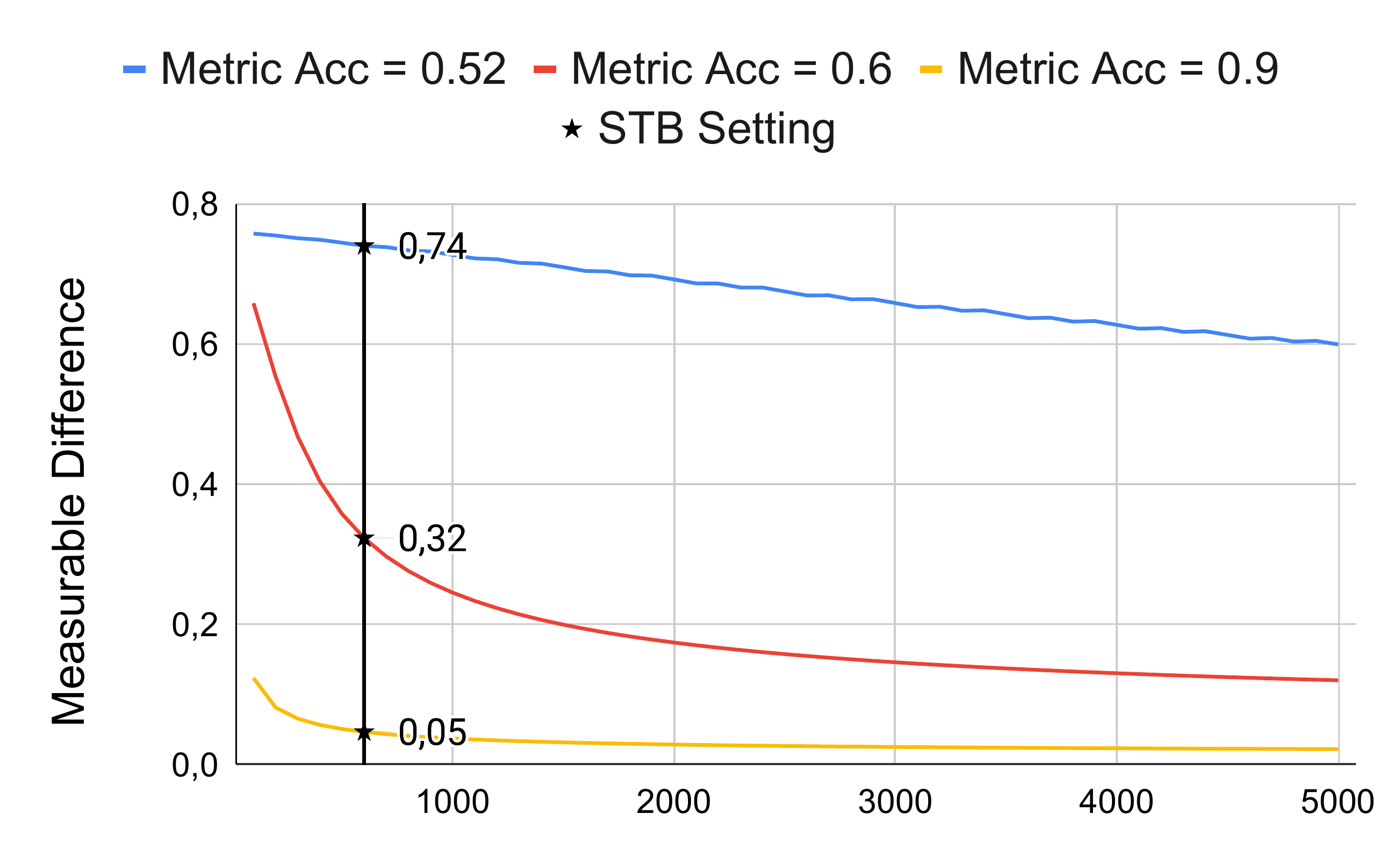}
         \caption{$|\mathcal{T}_{\Phi}| = 0$}
         \label{fig:stb_theory_a}
     \end{subfigure}
     \hfill
     \begin{subfigure}[b]{0.3\textwidth}
         \centering
         \includegraphics[width=1.2\textwidth]{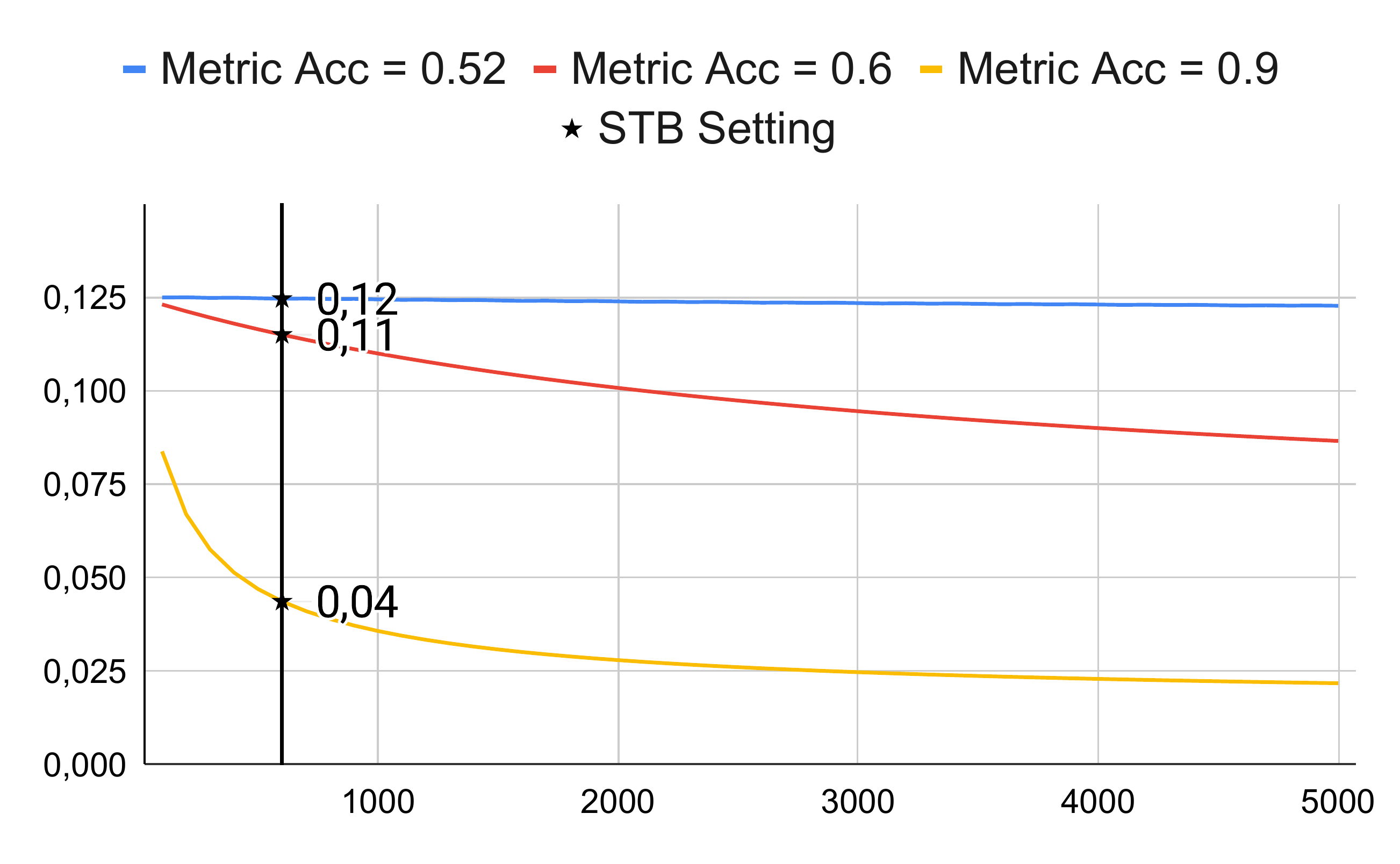}
         \caption{$|\mathcal{T}_{\Phi}| = 100$}
         \label{fig:stb_theory_b}
     \end{subfigure}
     \hfill
     \begin{subfigure}[b]{0.3\textwidth}
         \centering
         \includegraphics[width=1.2\textwidth]{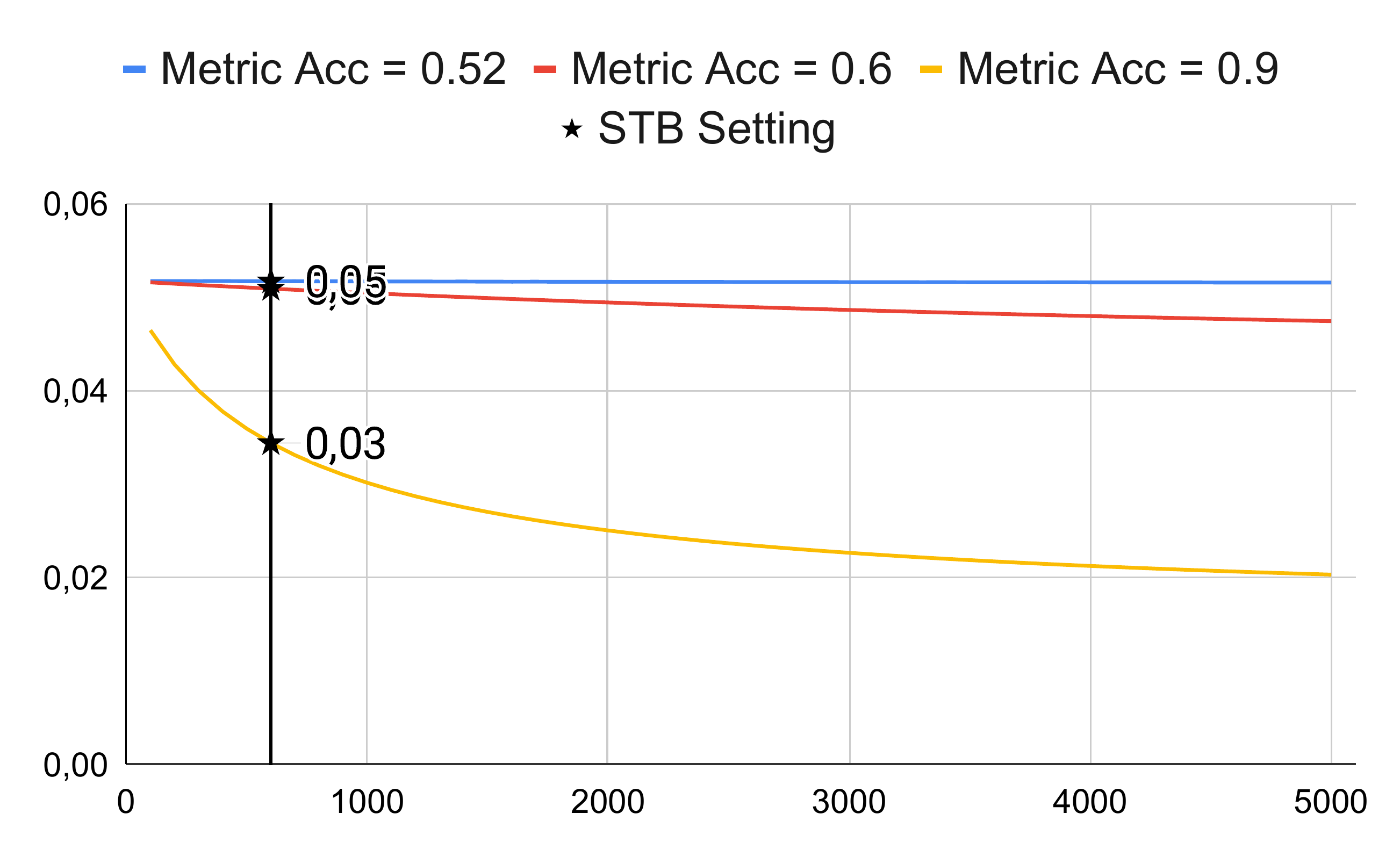}
         \caption{$|\mathcal{T}_{\Phi}| = 600$}
         \label{fig:stb_theory_c}
     \end{subfigure}
        \caption{Measurable difference (y-axis) depending on the number of samples available to estimate $\rho$ and $\eta$ (i.e, $|\mathcal{T}_{\rho, \eta}|$ on the x-axis). For fixed $|\mathcal{T}_M| = 10'000$. The vertical line denotes the STB setting with $|\mathcal{T}_{\rho, \eta}| = 600$}
        \label{fig:stb_theory}
\end{figure*}

\subsubsection{STB: Theoretical Bounds}
Figure~\ref{fig:stb_theory} shows the theoretical $\epsilon$ values that can be achieved depending on $|\mathcal{T}_{\rho, \eta}|$. The values are depicted for three different settings of $|\mathcal{T}_{\Phi}|$ (i.e, human ratings). Each setting shows the measurable $\epsilon$ for three different $\rho=\eta$ combinations. The figure reveals the impact of $|\mathcal{T}_{\rho, \eta}|$ for $|\mathcal{T}_{\rho, \eta}| < 1000$. For instance, for $|\mathcal{T}_{\rho, \eta}| = 600$, a metric with $\rho=\eta=0.6$ is only able to distinguish an $\epsilon = 0.11$, however, when increasing $|\mathcal{T}_{\rho, \eta}|$ to 5000 a difference of $\epsilon=0.08$ can be measured. On the other hand, when the performance of the metric is too low (e.g., $\rho=\eta=0.52$) the impact of higher $|\mathcal{T}_{\rho, \eta}|$ is negligible regardless of $|\mathcal{T}_{\Phi}|$. 


\subsubsection{STB: Practical Results}
\begin{table}[t!]
\centering
\small
\input{tables/showcase2/full_table}
\caption{Predicted STB evaluation using USR on three dialogue systems.}
\label{tbl:stb_pract_1} 
\end{table}

Table~\ref{tbl:stb_pract_1} shows the measured values for $\alpha$ and $\epsilon$ for three scenarios. The first two scenarios are analogous to the WMT setting, where we use $|\mathcal{T}_{\Phi}| = 600$ error-free ratings in the first scenario and $|\mathcal{T}_{\Phi}| = 100$ error-free ratings in the second scenario (assuming that we labeled only 100 samples due to cost reasons). For the third scenario we again use $|\mathcal{T}_{\Phi}| = 100$ error-free ratings, combined with $|\mathcal{T}_M| = 10'000$ error-prone ratings from the USR metric. The results show that for the first scenario all the pairs of systems are distinguishable, which is consistent with the theory and the original Spot The Bot results. When reducing the number of error-free samples to 
 $|\mathcal{T}_{\Phi}| = 100$, only the pair BL-KV is distinguishable. This is consistent with the theory, which predicts that two systems with $\epsilon > 0.126$ are significantly distinguishable. However, adding $|\mathcal{T}_M| = 10'000$ error-prone ratings only increases the probability of the first TG system being better than the second by a small amount. The reason is that the performance of USR is too low to have a strong impact, which is consistent with the theory. Thus, to benefit from automated evaluation one needs a better metric and more samples to estimate $\rho$ and $\eta$. 
 

\section{Related Work}
Evaluation of Text Generation systems is a long-standing issue. Considerations about the proper evaluation of TG systems have been emerging rapidly in the last years. One line of inquiry is how to properly conduct human evaluations and what kind of guidelines and setups lead to consistent results \cite{novikova2018rankme,van2019best,santhanam2019towards,freitag2021experts,clark-etal-2021-thats,belz2021reprogen,mohankumar2022active}. Another line of research investigates the reliability of automated metrics for NLG evaluation. \newcite[inter alia]{novikova-etal-2017-need} find that automated metrics poorly reflect human judgements in general. \newcite[Sec.\ 6]{sai2022survey} provides an extensive overview of criticism on automated metrics in NLG. 

There are few efforts to underlay (parts of the) TG evaluation paradigm with a theory-grounded base: To theoretically solidify human NLG evaluation and provide more statistically significant results in pairwise evaluations, a recent approach leverages utility theory in economics  \citep{ethayarajh2022human} to showcase issues arising from the use of Likert scale ratings and averaging them. \newcite{chaganty2018price} propose a method to combine automated metrics with human rankings to debiase a metric under a budget constraint. They provide a theory-grounded proof that their calculated mix of human and automated ratings is optimal and conclude that error-prone evaluation metrics are a bottleneck for reducing the cost of evaluations. Related to our Bayesian approach of modelling uncertainty in the evaluation of systems, a number of approaches aims to model uncertainty in the annotation process and the aggregation of annotations using a Bayesian approach \cite[e.g.]{paun2018comparing}. 
\citet{card-etal-2020-little} analyze the statistical power of different evaluation scenarios prevalent in NLP. In particular, they study the number of samples needed to detect a difference of 1 BLEU as significant. 
However, to the best of our knowledge, no efforts to model the uncertainties ingrained in TG evaluation in a holistic theory has been proposed so far.


\section{Conclusion}

We introduced a theoretical framework for binary metrics that can be used to extract guidelines for designing an evaluation of text generation systems. The framework estimates the performance of a text generation system from a mix of human and automated ratings giving guarantees of which level of significance can be achieved. Using the formulas, one can design the evaluation setup and compute estimates of how many human and automated samples are needed for a significant evaluation. We applied the theory to two very different real-world cases and exemplified how the theory can be leveraged to improve the significance of the results. We provide a tool that allows the computation of the formulas so that different settings can be tested.

The current theory is limited to binary metrics, but in future work, we will extend the theory to more types, such as comparative or scalar metrics. Furthermore, we will apply the theory to a wider range of tasks and domains. In general, we hope to have set in motion efforts to arrive at a sound formalization of the evaluation of text generation systems to increase the robustness, reliability, and significance of future evaluation campaigns.

\section*{Limitations}
\paragraph{Human Ratings.} We assume that human ratings are perfect, which is not the case~\cite{clark-etal-2021-thats}. While it might be the case that the MQM ratings are close to error-free, there is no guarantee. To handle the fact that human ratings are not error-free we would need to measure this, which could be done via agreement scores.

\paragraph{Uniform Input and Outputs.} We assume that each input and each output have the same difficulty of being evaluated. However, it is more likely that in practise, each metric has a different $\rho$ and $\eta$ value depending on the input. This is however very hard to include in the theory. 

\paragraph{Uniform Text Generation Systems.} Similarly to the above point, we assume that $\rho$ and $\eta$ are independent of the text generation system. However, preliminary experimental results (see Appendix \ref{app:roc}) showed that metrics tend to have different performances for different TG systems. Thus, $\rho$ and $\eta$ need to be estimated separately for each TG system. 

\paragraph{Domain Dependence.} The same argument can also be made about the domain. Metrics trained on one domain will perform differently when applied to another domain. Thus, the $\rho$ and $\eta$ values must be measured again for each domain. 

\paragraph{Binary Metrics.} The current theory is limited to binary metrics. However, in practise there are many different types of metrics and evaluation types. For instance, in a next step the theory should be extended to cover comparative metrics (i.e., metrics that state which of the two outputs is better).

\paragraph{Approximations.} The estimations of the mixed case and the estimated $\rho, \eta$ case must be approximated numerically since we did not find a closed form solution. This will inevitably lead to mistakes in the estimated values. This can be circumvented by making the numerical approximation more precise with the downside of needing more computational power (see Appendix~\ref{app:sim}).




\bibliography{anthology,custom}
\bibliographystyle{acl_natbib}

\appendix

\section{Derivations for $\alpha$-estimation}
\label{app:derivation}

In Section \ref{sec:theory} we have introduced several ways to estimate the
success rate $\alpha$ of a Text-Generator $\pi$. We will now elaborate some of these in more detail.

First, we want to estimate $\alpha$ based on consistent ratings from $M_{b}^{*}$. For this we need a set of inputs, the corresponding outputs from $\pi$, and the rating from $M_{b}^{*}$: $\mathcal{T}_{\Phi} = \{(i_j, o_j, r_{j}^{*}) | 1 \le j \le n_{\phi}\}$, where $o_j = \pi(i_j)$ and $r_{j}^{*} = M_{b}^{*}(i_j, o_j)$. We note that, in this case, the probability that a given pair is rated adequate is $\alpha$, since: 
\begin{equation*}
\begin{aligned}
     P(r_{j}^{*} = 1) &= P(M_{b}^{*}(i_j, \pi(i_j)) = 1) \\
     &= P(\pi(i_j) \in \mathcal{R}_{+}^{i_j}) \\
     &= \alpha
\end{aligned}
\end{equation*}
We can therefore treat $r_{j}^{*}$ as outcomes of Bernoulli trials with success probability $\alpha$. The number of successful trials $N_{+}$ is therefore a random variable with binomial distribution: $N_{+} \sim Binom(\alpha, n_{\phi})$. The concrete outcome for a given experiment is $n_{+} = \sum_{j = 1}^{n_{\phi}} r_{j}^{*}$. To estimate $\alpha$ we use the proportion of successful trials, meaning the fraction of adequate responses: $\tilde{\alpha} = \frac{n_+}{n_{\phi}}$. Due to the Law of Large Numbers this will converge to the expected value $\mathbb{E}[r_j] = \alpha$.

\paragraph{Bayesian Formulation} We choose to work in a Bayesian framework as it provides a convenient way to unify the multiple sources of evidence and uncertainty we want to tackle. The first source of information comes from $\mathcal{T}_{\Phi}$. In particular we have seen that the number of input-output pairs rated as adequate, $N_{+}$, follows a binomial distribution. This means that $P(N_{+} = n_{+} | \alpha) = \binom{n_{\phi}}{n_{+}} \alpha^{n_{+}}(1 - \alpha)^{n_{\phi} - n_{+}}$. We want to derive a posterior distribution for $\alpha$ based on the evidence: $p(\alpha | N_{+} = n_{+})$. For this we can apply Bayes' Theorem: $p(\alpha | N_{+} = n_{+}) \propto P(N_{+} = n_{+} | \alpha)p(\alpha)$, where $p(\alpha)$ expresses our prior belief of the possible values for $\alpha$. In this setting $p(\alpha)$ is called the prior, $P(N_{+} = n_{+} | \alpha)$ likelihood, and $p(\alpha | N_{+} = n_{+})$ the posterior.
Since we in general cannot assume anything about $\alpha$ we choose a uniform prior $\alpha \sim \mathcal{U}(0, 1)$. This means before seeing any evidence we consider any possible value of $\alpha$ to be equally likely. Of course there are other reasonable choices for priors, but in general uniform priors are a good choice since the resulting estimators will closely match traditional frequentist approaches.

Another approach is to choose a so-called conjugate prior based on the type of likelihood we are confronted with. A conjugate prior for a given likelihood will result in a posterior from the same family (but different parameters) as the prior. In our case, the Beta distribution is a conjugate prior for a Binomial likelihood. Beta distributions have two shape parameters $a$ and $b$ and assuming $\alpha \sim Beta(a, b)$ then $p(\alpha) = \frac{\alpha^{a - 1}(1 - \alpha)^{b - 1}}{B(a, b)}$. Here $B(a, b)$ is the beta function of $a$ and $b$ and serves as the normalizing constant, ensuring that $p(\alpha)$ integrates to $1$. The beta function is defined in terms of the Gamma function $\Gamma$, an extension of factorials.

Luckily, we can show that $\mathcal{U}(0, 1)$ and $Beta(1, 1)$ are the same distribution. We first note that both distributions are defined on the same domain $(0, 1)$. In particular, the uniform distribution is constant $1$ over the domain. By definition of the Beta distribution we have that if $\alpha \sim Beta(1, 1)$ then $p(\alpha) = \frac{\alpha^{1 - 1}(1 - \alpha)^{1 - 1}}{B(1, 1)} = \frac{1}{B(1, 1)} = 1$.

Next we will show how to compute the posterior for the general case where $\alpha \sim Beta(a, b)$:
\begin{equation*}
    \begin{aligned}
         &p(\alpha | N_{+} = n_{+})  \\
         &\propto P(N_{+} = n_{+} | \alpha)p(\alpha) \\
         &\propto \binom{n_{\phi}}{n_{+}}\alpha^{n_{+}}(1 - \alpha)^{n_{\phi} - n_{+}} \frac{\alpha^{a - 1}(1 - \alpha)^{b - 1})}{B(a, b)} \\
         &\propto \alpha^{n_{+} + a - 1}(1 - \alpha)^{n_{\phi} - n_{+} + b - 1} \\
         &\sim Beta(n_{+} + a, n_{\phi} - n_{+} + b)
    \end{aligned}
\end{equation*}
We see that the resulting posterior is indeed another Beta distribution. In particular if we choose $a = b = 1$, or a uniform prior, we get that $\alpha | N_{+} = n_{+} \sim Beta(n_{+} + 1, n_{\phi} - n_{+} + 1)$ as in Section \ref{sec:theory}.

\paragraph{Error-prone Metric} Next, we want to estimate $\alpha$ given a set of inputs, outputs from $\pi$, and ratings from a error-prone metric $M_{b}^{\rho, \eta}$ with known $\rho$ and $\eta$. We define $\mathcal{T}_{M} = \{(i_j, o_j, r_j) | 1 \le j < n_M \}$ where $o_j = \pi(i_j)$ and $r_j = M_{b}^{\rho, \eta}(i_j, o_j)$. 
The probability that any given $r_j$ is $1$ is:
\begin{equation*}
    \begin{aligned}
    &P(r_j = 1) \\
    &= P(M_{b}^{\rho, \eta}(i_j, \pi(i_j) = 1) \\
    &= P(M_{b}^{\rho, \eta}(i_j, \pi(i_j)) = 1 | \pi(i_j) \in \mathcal{R}_{+}^{i_j}) \\
    &P(\pi(i_j) \in \mathcal{R}_{+}^{i_j}) \\
    &+ P(M_{b}^{\rho, \eta}(i_j, \pi(i_j)) = 1 | \pi(i_j) \notin \mathcal{R}_{+}^{i_j}) \\
    &P(\pi(i_j) \notin \mathcal{R}_{+}^{i_j}) \\
    &=\rho\alpha \\
    &+ (1 - P(M_{b}^{\rho, \eta}(i_j, \pi(i_j)) = 0 | \pi(i_j) \notin R_{+}^{i_j})) \\
    &(1 - P(\pi(i_j) \in R_{+}^{i_j})) \\
    &=\rho\alpha + (1 - \eta)(1 - \alpha) \\
    &= \alpha(\rho + \eta - 1) + (1 - \eta) \\
    \end{aligned}
\end{equation*}

What we can concretely measure (or count) on $\mathcal{T}_{M}$ is the number of times the error-prone metric gives an adequate rating. We define this as $m_{+} = \sum_{j = 1}^{n_M} r_j$. Since we sum $n_M$ Bernoulli trials with success rate $\alpha(\rho + \eta - 1) + (1 - \eta)$, the sum has a Binomial distribution: $M_{+} \sim Binom(\alpha(\rho + \eta - 1) + (1 - \eta), n_M)$. Therefore our likelihood is:
\begin{equation*}
    \begin{aligned}
    &P(M_{+} = m_{+} | \alpha, \rho, \eta) \\
    &= \binom{n_M}{m_{+}}(\alpha(\rho+\eta-1)+(1-\eta))^{m_{+}} \\
    &(1 - (\alpha(\rho+\eta-1)+(1 - \eta)))^{n_M - m_+}
    \end{aligned}
\end{equation*}
We notate the likelihood as $P(M_{+} = m_+ | \alpha, \rho, \eta)$ to indicate the dependence on $\rho$ and $\eta$, even though they are assumed deterministic. Unfortunately, we are not aware of any conjugate prior for $\alpha$ that would allow us to derive a closed form posterior from this likelihood. Nevertheless, we can show that for $\alpha \sim \mathcal{U}(0, 1)$ the mode of the posterior is at $\frac{\frac{m_+}{n_M}- (1 - \eta)}{\rho + \eta - 1}$. For this we will have to find the point where the derivative of the posterior with respect to $\alpha$ is $0$.
To simplify the notation we will write $f(\alpha) = \alpha(\rho+\eta-1) + (1 - \eta)$ and $f'(\alpha) = \frac{\mathrm{d}}{\mathrm{d}\alpha}f(\alpha) = \rho + \eta - 1$.

We will first compute the derivative of the posterior with respect to $\alpha$ using a uniform prior (i.e. $p(\alpha) = 1$):
\begin{equation*}
    \begin{aligned}
    &\frac{\mathrm{d}}{\mathrm{d}\alpha}p(\alpha | M_+ = m_+) \\
    &\propto \frac{\mathrm{d}}{\mathrm{d}\alpha}P(M_+ = M_+ | \alpha, \rho, \eta)p(\alpha) \\
    &\propto \frac{\mathrm{d}}{\mathrm{d}\alpha}P(M_+ = M_+ | \alpha, \rho, \eta)1 \\
    &\propto \frac{\mathrm{d}}{\mathrm{d}\alpha} (f(\alpha)^{m_+}(1 - f(\alpha))^{n_M - m_+}) \\
    &\propto (\frac{\mathrm{d}}{\mathrm{d}\alpha}f(\alpha)^{m_+})(1 - f(\alpha))^{n_M - m_+} \\
    &+ f(\alpha)^{m_+}(\frac{\mathrm{d}}{\mathrm{d}\alpha}(1 - f(\alpha))^{n_M - m_+})) \\
    &\propto m_+f(\alpha)^{m_+ - 1}f'(\alpha)(1 - f(\alpha))^{n_M - m_+} \\
    &+ f(\alpha)^{m_+}(n_M - m_+)(1 - f(\alpha))^{n_M - m_+ - 1}
    (- f'(\alpha)) \\
    &\propto m_+f'(\alpha)f(\alpha)^{m_+ - 1}(1 - f(\alpha))^{n_M - m_+} \\
    &- (n_M - m_+)f'(\alpha)f(\alpha)^{m_+}(1 - f(\alpha))^{n_M - m_+ - 1}\\
    \end{aligned}
\end{equation*}

To find the mode we set the derivative to zero and solve for $\alpha$. We will use the convenient fact that $f'(\alpha)$ is constant independent of $\alpha$:
\begin{equation*}
    \begin{aligned}
    &m_+f'(\alpha)f(\alpha)^{m_+ - 1}(1 - f(\alpha))^{n_M - m_+} \\
    &- (n_M - m_+)f'(\alpha)f(\alpha)^{m_+}(1 - f(\alpha))^{n_M - m_+ - 1} \\
    &= 0 \iff \\
    &m_+f'(\alpha)f(\alpha)^{m_+ - 1}(1 - f(\alpha))^{n_M - m_+} \\
    &= (n_M - m_+)f'(\alpha)f(\alpha)^{m_+}(1 - f(\alpha))^{n_M - m_+ - 1} \\
    &m_+f(\alpha)^{m_+ - 1}(1 - f(\alpha))^{n_M - m_+}\\
    &=(n_M - m_+)f(\alpha)^{m_+}(1 - f(\alpha))^{n_M - m_+ - 1}\\
    &m_+(1 - f(\alpha))^{n_M - m_+}\\
    &=(n_M - m_+)f(\alpha)(1 - f(\alpha))^{n_M - m_+ - 1}\\
    & m_+(1 - f(\alpha)) = (n_M - m_+)f(\alpha) \\
    & m_+ = (n_M - m_+)f(\alpha) + m_+f(\alpha) \\
    & m_+ = (n_M - m_+ + m_+)f(\alpha) \\
    & \frac{m_+}{n_M} = f(\alpha) = \alpha(\rho+\eta-1)+(1 - \eta)\\
    & \alpha = \frac{\frac{m_+}{n_M} - (1 - \eta)}{\rho + \eta - 1}
    \end{aligned}
\end{equation*}

\paragraph{Uncertainty in $\rho$ and $\eta$} If we do not already know the specific $\rho$ and $\eta$ for a given error-prone metric, we will have to estimate them from data. For this we need ratings from a the error-prone metric as well as an error-free metric to compare to. Assume we are given the set $\mathcal{T}_{\rho, \eta} = \{(i_j, o_j, r_j, r_j^*) | 1 \le j < n_{\rho, \eta}\}$, where $r_j = M_{b}^{\rho, \eta}(i_j, o_j)$ and $r_j^* = M_b^*(i_j, o_j)$. Note that unlike $\mathcal{T}_{\Phi}$ and $\mathcal{T}_M$ we do not make any assumptions about how $o_j$ was generated.

By definition $\rho$ is the true positive rate of the error-prone metric and $\eta$ the true negative rate. We can therefore estimate them independently from each other by splitting $\mathcal{T}_{\rho, \eta}$ into two sets based on whether $r_j^*$ is $1$ or $0$:
$\mathcal{T}_{\rho, \eta}^{+} = \{(i, o, r, r^*) \in \mathcal{T}_{\rho, \eta}| r^* = 1\}$ and $\mathcal{T}_{\rho, \eta}^{-} = \{(i, o, r, r^*) \in \mathcal{T}_{\rho, \eta}| r^* = 0\}$.

To estimate $\rho$ we have to count the number of times $r_j = 1$ when $r_j^* = 1$ too, in other words we have to count the number of true adequate ratings: $n_{TP} = \sum_{i, o, r, r* \in \mathcal{T}_{\rho, \eta}^+} r$. By definition we know that $\rho = P(r = 1 | r* = 1)$ and therefore $N_{TP} \sim Binom(\rho, |\mathcal{T}_{\rho, \eta}^{+}|)$. We can apply the same Bayesian reasoning as at the start of this Appendix to derive a posterior distribution for $\rho$. Assuming a uniform prior over $\rho$, we have that $\rho | n_{TP} \sim Beta(n_{TP} + 1, |\mathcal{T}_{\rho, \eta}^{+}| - n_{TP} + 1)$. The estimation of $\eta$ is exactly analogous.

At this point we could just use point estimates for $\rho$ and $\eta$ and treat them as deterministic like above. Unfortunately this has a high chance of throwing off the point estimate (mode) of $\alpha$.

We will therefore consider the joint likelihood $P(M_{+} = m_{+}, \rho, \eta | \alpha)$ and marginalize $\rho$ and $\eta$. We will reuse results from above. Recall we were given the set $\mathcal{T}_{M} = \{(i_j, o_j, r_j) | 1 \le j < n_M \}$ where $o_j = \pi(i_j)$ and $r_j = M_{b}^{\rho, \eta}(i_j, o_j)$. We counted the number of adequate ratings $m_+ = \sum_{j = 1}^{n_M} r_j$ and we saw that $P(M_+ = M_+ | \alpha, \rho, \eta) = \alpha(\rho+\eta-1)+(1 - \eta)$. Based on that we can compute the likelihood as follows:
\begin{equation*}
    \begin{aligned}
    &P(M_+ = m_+ | \alpha) \\
    &= \int_0^1\int_0^1P(M_+ = m_+, \rho, \eta | \alpha)\mathrm{d}\rho\mathrm{d}\eta \\
    &= \int_0^1\int_0^1P(M_+ = m_+ | \rho, \eta, \alpha)p(\rho)p(\eta)\mathrm{d}\rho\mathrm{d}\eta \\
    \end{aligned}
\end{equation*}
and the posterior as follows:
\begin{equation*}
    \begin{aligned}
    &p(\alpha | M_+ = m_+) \\
    &\propto p(\alpha)P(M_+ = m_+ | \alpha) \\
    &\propto p(\alpha)\int_0^1\int_0^1P(M_+ = m_+ | \rho, \eta, \alpha)p(\rho)p(\eta)\mathrm{d}\rho\mathrm{d}\eta \\
    \end{aligned}
\end{equation*}

We will show how approximate this numerically in Appendix \ref{app:sim}.

\paragraph{Combining error-free and error-prone ratings} Finally, we show how we can combine both error-free and error-prone ratings into a single estimate for $\alpha$. Here we assume that we have estimates for $\rho$ and $\eta$, for example in the form of Beta-posteriors, as derived previously: $\rho \sim Beta(a_{\rho}, b_{\rho})$ and $\eta \sim Beta(a_{\eta}, b_{\eta})$. Similarly, we build upon the previous setting where we counted the number of adequate ratings from the error-free metric, $N_{+} \sim Binom(\alpha, n_{\phi})$, and the number of adequate ratings from the error-prone metric, $M_{+} \sim Binom(\alpha(\rho+\eta-1) + (1 - \eta), n_M)$. Our observed $n_+$ and $m_+$ have the joint likelihood:
\begin{equation*}
    \begin{aligned}
    &P(M_{+} = m_+, N_+ = n_+ | \alpha) \\
    &= P(M_+ = m_+ | \alpha)P(N_+ = n_+ | \alpha)
    \end{aligned}
\end{equation*}
We assume here that $M_+$ and $N_+$ are independent when conditioned on $\alpha$.

We are now ready to compute the posterior for $\alpha$. Using a Beta prior $\alpha \sim Beta(a_{\alpha}, b_{\alpha})$ we get:
\begin{equation*}
    \begin{aligned}
    &p(\alpha | M_+ = m_+, N_+ = n_+) \\
    &\propto p(\alpha)P(M_+ = m_+, N_+ = n_+ | \alpha) \\
    &\propto p(\alpha)P(N_+ = n_+ | \alpha)P(M_+ = m_+ | \alpha) \\
    &\propto p(\alpha)P(N_+ = n_+ | \alpha)\\
    &\int_0^1\int_0^1
    P(M_+ = m_+ | \alpha, \rho, \eta)p(\rho)p(\eta)
    \mathrm{d}\rho\mathrm{d}\eta \\
    &\propto p(\alpha | N_+ = n_+) \\
    &\int_0^1\int_0^1P(M_+ = m_+ | \alpha, \rho, \eta)p(\rho)p(\eta)
    \mathrm{d}\rho\mathrm{d}\eta \\
    \end{aligned}
\end{equation*}

Looking at the last step, we see that we can combine the prior $p(\alpha)$ with the partial likelihood $P(N_+ = n_+ | \alpha)$ to get a partial posterior $p(\alpha | N_+ = n_+)$ that gets multiplied with the likelihood of $M_+$. We have already seen that since $\alpha$ has a Beta prior and $N_+$ has a binomial likelihood, $\alpha | N_+$ is also a Beta distribution. This suggests a two-step procedure, where in the first step we derive a posterior from error-free ratings. In the second step we use that estimate as the new prior for deriving the posterior from error-prone ratings.

\paragraph{Notes on $\mathcal{T}_{\Phi}$, $\mathcal{T}_M$, and $\mathcal{T}_{\rho, \eta}$}
Note that in practise there are some considerations to be made. Since we use human ratings, we can use them both for estimating $\rho$ and $\eta$ but also to estimate $\alpha$. Thus, we use $T_{\Phi} = T_{\rho, \eta}$, which is also necessary since $\rho$ and $\eta$ are different for each TG system (see example in Appendix~\ref{app:roc}). Thus, it is often not advisable to use the ratings for other systems to estimate $\rho$ and $\eta$. However, this phenomenon needs to be explored in more detail.

For the estimation of $\rho$ and $\eta$, we need to make sure that $T_{\rho, \eta}^+$ and $T_{\rho, \eta}^-$ are of large enough size. Since if we have only a few samples in $T_{\rho, \eta}$ where $r_j^*=0$ then the estimate for $\eta$ will be uncertain. This can be problematic when evaluating very strong or very poor systems (e.g., $\alpha > 0.9$ or $\alpha < 0.1$) as there will be only a few samples with $r_j^*=0$ or $r_j^*=1$ respectively. 

In many cases we can reuse the samples in $T_{\Phi}$ for $T_M$, i.e., $T_{\Phi} \subseteq T_M$ since we can use the automated metric to rate the samples, which were annotated by humans. However, it is not clear what effect this will have on the final estimate of $\epsilon$. Exploring this phenomenon is part of future work. 

\section{Derivations for $\epsilon$-simulation}
\label{app:sim}

In this section we will show how we derive the values for the minimally distinguishable difference between two systems. We do this by first simulating a concrete experiment based on theoretical parameters. We substitute the simulated experiment into Equation \ref{eq:fusion}. We will also show how we numerically approximate Equation \ref{eq:fusion}.

\paragraph{Simulation} Until now we have considered the case where $\alpha$ , and possibly $\rho$ and $\eta$, are unknown and need to be estimated from data. In that case we use Equation \ref{eq:fusion} to derive a posterior estimate for $\alpha$. The whole estimation is based on counts from three sources $\mathcal{T}_{\Phi}$, $\mathcal{T}_M$, and $\mathcal{T}_{\rho, \eta}$. Assume we know the following properties: $\alpha$, $\rho$, $\eta$, $n_{\phi} = |\mathcal{T}_{\Phi}|$, $n_M = |\mathcal{T}_M|$, $n_{\rho, \eta} = |\mathcal{T}_{\rho, \eta}|$, as well as the proportion $\psi$ of truly adequate responses in $\mathcal{T}_{\rho, \eta}$. 

To simulate the number of adequate ratings from the error-free metric $n_+$ we round its expected value, $\mathbb{E}[n_+] = \alpha n_{\phi}$, to the nearest integer: $n_{+}^{sim} = \lfloor \alpha n_{\phi} + \frac{1}{2} \rfloor$. To simulate the number of adequate ratings from the error-prone metric $m_+$, we round its expected value, $\mathbb{E}[m_+] = (\alpha(\rho + \eta - 1) + (1 - \eta))n_M$, to the nearest integer: $m_{+}^{sim} = \lfloor (\alpha(\rho + \eta - 1) + (1 - \eta))n_M + \frac{1}{2} \rfloor$.
We have seen that to estimate $\rho$ we need to know the number of true positive ratings $n_{TP}$ of the error-prone metric as well as the total number of positive ratings in $\mathcal{T}_{\rho, \eta}$ which we notated as $|\mathcal{T}_{\rho, \eta}^{+}| = n_p^*$. We can simulate the latter by rounding its expected value, $\mathbb{E}[n_p^*] = \psi n_{\rho, \eta}$, to the nearest integer: $n_p^{sim} = \lfloor \psi n_{\rho, \eta} + \frac{1}{2} \rfloor$. To simulate $n_{TP}$ we have to plug the simulated $n_p^{sim}$ into the expected value: $n_{TP}^{sim} = \lfloor \rho n_p^{sim} + \frac{1}{2} \rfloor$. Finally, we follow the same process to simulate the data for $\eta$. Let $n_n^* = |\mathcal{T}_{\rho, \eta}^{-}|$, which we simulate as $n_n^{sim} = n_{\rho, \phi} - n_p^{sim}$. The number of true negatives of the error-prone metric is simulated as: $n_{TN}^{sim} = \lfloor \eta(n_{\rho, \phi} - n_p^{sim}) + \frac{1}{2} \rfloor$.

We can then use these simulated values to calculate the calculate the posterior $p^{sim}(\alpha)$ based on Equation \ref{eq:fusion}. For this we first have to simulate our belief over $\rho$ and $\eta$: $\rho^{sim} \sim Beta(n_{TP}^{sim} + 1, n_p^{sim} - n_{TP}^{sim} + 1)$ and $\eta^{sim} \sim Beta(n_{TN}^{sim} + 1, n_n^{sim} - n_{TN}^{sim} + 1)$. We again set a uniform prior, $\alpha \sim Beta(1, 1)$ and compute the simulated posterior:
\begin{equation*}
    \begin{aligned}
    &p^{sim}(\alpha | N_+ = n_+^{sim}, M_+ = n_+^{sim}) \\
    &\propto p(\alpha)P(N_+ = n_+^{sim} | \alpha) \\
    &\int_0^1\int_0^1P(M_+ = m_+^{sim} | \alpha, \rho, \eta)p^{sim}(\rho)p^{sim}(\eta)\mathrm{d}\rho\mathrm{d}\eta
    \end{aligned}
\end{equation*}

For the tables in Appendix \ref{app:tab_theory} we make the following simplifying assumptions: we assume that the input-output pairs in $\mathcal{T}_{\Phi}$ and $\mathcal{T}_{\rho, \eta}$ are the same. This means that $n_{\rho, \eta} = n_{\phi}$ and $\psi = \alpha$.

\paragraph{Computing $\epsilon_{\gamma}$} We will now show how we use $p^{sim}(\alpha)$ to compute the minimal distinguishable difference between two systems $\pi_1$ with success rate $\alpha_1$ and $\pi_2$ with success rate $\alpha_2$. 

Assume we know the distributions $p(\alpha_1)$ and $p(\alpha_2)$, we can then compute their means $\mu_i = \mathbb{E}[\alpha_i]$ and variances $\sigma_i^2 = \mathbb{V}[\alpha_i]$. These can be used to derive normal approximations for $\alpha_i$: $\alpha_i^{\mathcal{N}} \sim \mathcal{N}(\mu_i, \sigma_i)$. In that case the difference $\epsilon = \alpha_1^{\mathcal{N}} - \alpha_2^{\mathcal{N}}$ also follows a normal distribution: $\mathcal{N}(\mu_1 - \mu_2, \sqrt{\sigma_1^2 + \sigma_1^2})$.
We can now formulate a z-test to see whether there is a significant difference between $\alpha_1^{\mathcal{N}}$ and $\alpha_1^{\mathcal{N}}$. The null hypothesis $H_0$ is that both systems perform the same, meaning $\mu_1 = \mu_2$ or $\epsilon = 0$. Under $H_0$ we have that $\frac{\epsilon}{\sqrt{\sigma_1^2 + \sigma_1^2}} \sim \mathcal{N}(0, 1)$. To reject $H_0$ at a certain significance level $\gamma$, we have to show that $|\frac{\epsilon}{\sqrt{\sigma_1^2 + \sigma_1^2}}| > \Phi^{-1}(1 - \frac{\gamma}{2})$. Here $\Phi^{-1}$ is the inverse cumulative distribution function of the standard normal distribution and we notate $Z_{\gamma} = \Phi^{-1}(1 - \frac{\gamma}{2})$. In that case, all $|\epsilon| > \sqrt{\sigma_1^2 + \sigma_2^{2}}Z_{\gamma}$ will be significant under this test. The minimal significant difference at the $\gamma$ level is then $\epsilon_{\gamma} = \sqrt{\sigma_1^2 + \sigma_2^{2}}Z_{\gamma}$.

Given our simulated posterior $p^{sim}(\alpha)$ we can compute its mean, $\mu^{sim} = \int_0^1 \alpha p^{sim}(\alpha)\mathrm{d}\alpha$ and variance $\sigma_{sim}^{2} = \int_0^1 (\alpha - \mu^{sim})^2 p^{sim}(\alpha) \mathrm{d}\alpha$. We have to make one final assumption: if we estimate $\alpha_1$ and $\alpha_2$ under exactly the same conditions, meaning with the same $n_{\phi}$, $n_M$, $n_{\rho, \eta}$ and the same error-prone metric $M_{b}^{\rho, \eta}$, and their difference $\epsilon$ is relatively small, then their variances should be the same. Using this assumption we compute: $\epsilon_{\gamma}^{sim} = \sqrt{2\sigma_{sim}^{2}}Z_{\gamma}$.

\paragraph{Caveats} At this point we will reflect on the several layers of approximations we go through to arrive at an numerical estimate for $\epsilon_{\gamma}$. We start out by simulating an experiment where we replace all key observables by their expected values under our experiment assumptions (i.e. the chosen fixed values of $\alpha$, $\rho$, $\eta$ and sample sizes). Of course in a real world setting those values could deviate from their expected values due to bad luck. This will influence both the mean and variance of the resulting estimate. We then compute the simulated posterior using numerical approximation (see next paragraph), which could be imprecise. We then further approximate the posterior by a normal distribution. In practice, we work with large enough sample sizes, that the normal approximation should be relatively accurate.

The overall implication is that the theoretical values of $\epsilon_{\gamma}$ we use throughout this work provide a useful guideline but it is unclear how exact they are.

\paragraph{Numerical Approximation of Posteriors} A problem we face repeatedly is that we are interested in the expected values of a function of a continuous random variable, such as $\int_a^b f(x)p(x)\mathrm{d}x$, which might not have an easily computable closed form. This is for example the case for the integrals over $\rho$ and $\eta$ in Equation \ref{eq:fusion}, but also when computing the mean and variance of the posterior.

We will now elaborate how we approximate expected values of a continuous variable by middle Riemann sums. Assume we are given a random variable $x$ with domain $(0, 1)$, its density function $p(x)$, and its cumulative density function $CDF_x(x') = P(x < x') = \sum_0^{x'}p(x)\mathrm{d}x$. The main idea is to partition the domain into a discrete number equally sized slices. Every partition gets identified by its midpoint and the total total density within that partition. Let $N_x$ be the number of slices, the larger $N_x$ the preciser our approximation will be. We define:
\begin{equation*}
    \begin{aligned}
    \forall 0 \le i < N_x \\
    \boldsymbol{x}[i] &= \frac{i + \frac{1}{2}}{N_x} \\
    \boldsymbol{P}_x[i] &= \int_{\frac{i}{N_x}}^{\frac{i + 1}{N_x}}p(x)d(x) \\
    &= \int_{0}^{\frac{i + 1}{N_x}}p(x)d(x) - \int_{0}^{\frac{i}{N_x}}p(x)d(x) \\
    &= CDF_x(\frac{i+1}{N_x}) - CDF_x(\frac{i}{N_x}) \\
    \end{aligned}
\end{equation*}
Here $\boldsymbol{x}[i]$ represents the midpoint of the interval $[\frac{i}{N_x}, \frac{i+1}{N_x}]$ and $\boldsymbol{P}_x[i]$ the total probability mass in that interval. To approximate expected values we can now replace integrals by sums: $\mathbb{E}[f(x)] = \int_0^1f(x)p(x)\mathrm{d}x \approx \sum_{i=0}^{N_x - 1}f(\boldsymbol{x}[i])\boldsymbol{P}_x[i]$.

If we want to apply this discretization to $\alpha$, $\rho$, and $\eta$, we need access to their cumulative distribution functions. In our framework, these variables are either uniformly or more generally Beta distributed. The cumulative distributions for these are available in most numerical software libraries and therefore computing the discretization is relatively straight-forward.

Applying this discretization to $\alpha$, $\rho$, and $\eta$ we can restate Equation \ref{eq:fusion} in approximate form:
\begin{equation}\label{eq:fusion_approx}
    \begin{aligned}
    &\boldsymbol{P}_{\alpha | N_+ = n_+, M_+ = m_+}[i] =\\
    &\boldsymbol{P}_{\alpha}[i]P(N_+ = N_+ | \boldsymbol{\alpha}[i]) \\
    &\sum_{j = 0}^{N_{\rho} - 1} \sum_{k = 0}^{N_{\eta} - 1} \bigg(
    P(M_+ = m_+ | \boldsymbol{\alpha}[i], \boldsymbol{\rho}[j], \boldsymbol{\eta}[k]) \\
    &\boldsymbol{P}_{\rho}[j] \boldsymbol{P}_{\eta}[k] \bigg) \\
    &\forall 0 \le i < N_{\alpha}
    \end{aligned}
\end{equation}

This results in a discretized form of the posterior with the same granularity $N_{\alpha}$ as for the prior. We can then approximate the mean of the posterior as:
\begin{equation*}
    \begin{aligned}
    &\mathbb{E}[\alpha | N_+ = n_+, M_+ = m_+] \approx \\
    &\sum_{i = 0}^{N_{\alpha}} 
    \boldsymbol{P}_{\alpha | N_+ = n_+, M_+ = m_+}[i]
    \boldsymbol{\alpha}[i]
    \end{aligned}
\end{equation*}
and the variance as:
\begin{equation*}
    \begin{aligned}
    &\mathbb{V}[\alpha | N_+ = n_+, M_+ = m_+] \\
    &=\mathbb{E}[\alpha^2|N_+=n_+,M_+ = m_+] \\
    &- (\mathbb{E}[\alpha|N_+=n_+,M_+=m_+])^2 \\
    &\approx \sum_{i = 0}^{N_{\alpha} - 1} \boldsymbol{P}_{\alpha | N_+ = n_+, M_+ = m_+}[i]\boldsymbol{\alpha}^{2}[i] \\
    &- \bigg(\sum_{i = 0}^{N_{\alpha} - 1} \boldsymbol{P}_{\alpha | N_+ = n_+, M_+ = m_+}[i]\boldsymbol{\alpha}[i]\bigg)^{2} \\
    \end{aligned}
\end{equation*}

We use $N_{\alpha} = 2000$ and $N_{\rho} = N_{\eta} = 1000$ in all our experiments.

\section{ROC Curves of Metrics}\label{app:roc}

\begin{figure}
    \centering
    \includegraphics[width=\columnwidth]{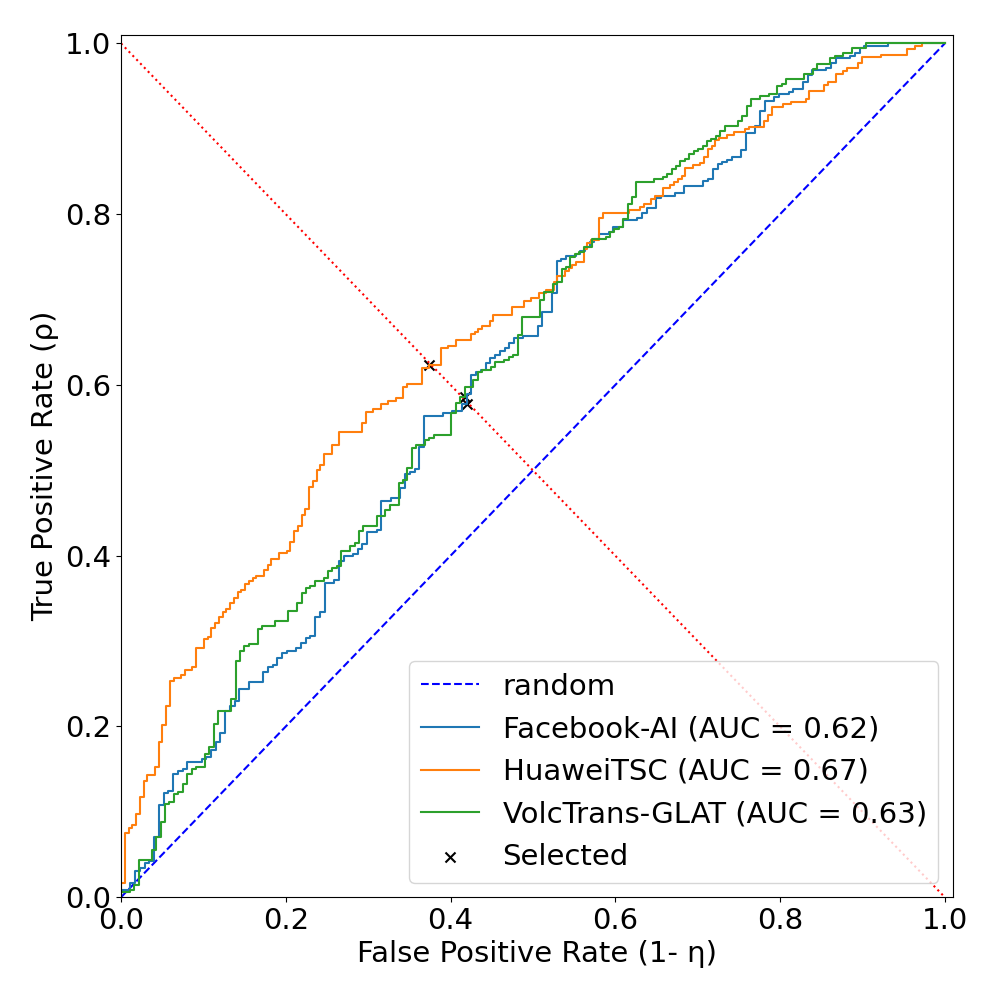}
    \caption{ROC curves for BleuRT predicting MQM annotations for 3 MT system. The markers show the threshold we select in our experiments. The blue diagonal corresponds to a random baseline. The red diagonal visualizes points where $\rho = \eta$.}
    \label{fig:bleurt_roc}
\end{figure}

While our theory assumes binary metrics that will only produce $0$ or $1$ ratings, most real-world automated metrics produce scalar ratings $\in \mathbb{R}$. In our case all metrics under consideration produce scalar ratings. To apply our framework we have to transform scalar ratings into binary ratings. We can do this by selecting a threshold $\tau$ that partitions the ratings into binary classes (on either side of the threshold). We define a scalar metric as a function of input output pairs to the reals: $M_{s}: \mathcal{I} \times \mathcal{O} \rightarrow \mathbb{R}$. We interpret the rating as a preference, such that, if $M_s(i, o_1) > M_s(i, o_2)$, then we say that according to $M_s$, $o_1$ fits $i$ better than $o_2$. Given a scalar metric $M_s$ and a threshold $\tau \in \mathbb{R}$ we can derive the associated binary metric:
\begin{equation}\label{eq:binarization}
    M_{b}^{\tau}(i, o) =
    \begin{cases}
    1 & \text{if } M_s(i, o) > \tau \\
    0 & \text{else}
    \end{cases}
\end{equation}

The question is now, how to select $\tau$. This is a well known problem in binary classification. Intuitively, every possible threshold $\tau$ is associated with a pair of corresponding $\rho$ and $\eta$.

Figure \ref{fig:bleurt_roc} shows the Receiver Operator Characteristic (ROC) curves for BleuRT as a predictor of $M_{b}^{*}$ for three machine translation systems. In an ROC plot, the true positive rate is plotted against the false positive rate at various thresholds $\tau$. We note that in our framework the true positive rate is $\rho$ and the false positive rate is $1 - \eta$. 

Assume we are given a set of inputs and outputs, the ratings from $M_s$ and ratings from an error-free binary metric $M_b^*$:
$\mathcal{T}_S = \{(i_j, o_j, s_j, r_j^*) | 1 \le j \le n_S\}$, where $s_j = M_s(i_j, o_j)$ and $r_j^* = M_{b}^{*}(i_j, o_j)$. We can consider the values $s_j$ as candidate thresholds, as these are exactly the cases where the predictions would switch in Equation \ref{eq:binarization}. For each candidate threshold, we can binarize the predictions and compute the associated $\rho$ and $\eta$. We select the threshold that minimizes $|\rho - \eta|$, to be consistent with our examples, where we usually assumed for simplicity that $\rho = \eta$. This selection is shown in Figure \ref{fig:bleurt_roc} by markers and the red diagonal.

One thing to note in Figure \ref{fig:bleurt_roc} is that the curves for the three MT systems differ from each other. This means that the specific $\rho$ and $\eta$ of BleuRT when used as a binary metric depend on the systems that produced a given output. In our framework laid out in Section \ref{sec:theory} we assumed that $\rho$ and $\eta$ are independent of how a given output $o$ is produced. This calls for further analysis in future work.

\section{Full Show Cases Tables} \label{app:sc_tbl}
In this Appendix, we show the full tables for the show cases with all the systems from the WMT and STB setting. 
\subsection{WMT21}
For the WMT task, we have 4 scenarios (see Section~\ref{sec:wmt_sc}), for all these scenarios we show the pariwise comparisons in Tables~\ref{tab:wmt21_app1}, \ref{tab:wmt21_app2}, \ref{tab:wmt21_app3}, and \ref{tab:wmt21_app4}. Each table shows for each system the estimated $\alpha$ value in parentheses, and in each cell the $\epsilon$ value with $P(\alpha_1 > \alpha_2)$ in parentheses. All the pairs that are significantly distinguishable are put in bold. The WMT systems are Facebook-AI (FBAI)~\cite{tran-etal-2021-facebook}, VolcTrans-GLAT (VT-G)~\cite{qian-etal-2021-volctrans}, Online-W (OW)~\footnote{Anonymous System}~\cite{}, Nemo (NE), VolcTrans-AT (VT-A)~\cite{qian-etal-2021-volctrans}, UEdin (UE)~\cite{chen-etal-2021-university}, and HuaweiTSC (HU)~\cite{wei-etal-2021-hw}. Table~\ref{tab:wmt21_app1} where all the human ratings are used shows that FBAI, VT-G, OW, NE, and VT-A are not significantly distinguishable from eachother as their $\epsilon < 0.06$. For the other three scenarios none of the systems are distinguishable. This is consistent with the theoretical predictions. From Table~\ref{tbl:intro_a}, we see that at least 5000 human ratings are needed to be able to significantly distinguish all the pairs of systems (i.e., for $\epsilon < 0.02$). Thus, in this case the problem is that the TG systems are too close to eachother in terms of performance and the automated metrics are too weak to boost the evaluation with low cost. 
\begin{table*}[]
    \centering
    \small
    \input{tables/appendixD/wmt_to527_tm0}
    \caption{Full WMT scenario with $|\mathcal{T}_{\Phi}| = 527$, and $|\mathcal{T}_M| = 0$}
    \label{tab:wmt21_app1}
\end{table*}

\begin{table*}[]
    \centering
    \small
    \input{tables/appendixD/wmt_to100_tm0}
    \caption{Full WMT scenario with $|\mathcal{T}_{\Phi}| = 100$, and $|\mathcal{T}_M| = 0$}
    \label{tab:wmt21_app2}
\end{table*}

\begin{table*}[]
    \centering
    \small
    \input{tables/appendixD/wmt_to100_tm1000_sentbleu}
    \caption{Full WMT scenario with $|\mathcal{T}_{\Phi}| = 100$, and $|\mathcal{T}_M| = 1000$}
    \label{tab:wmt21_app3}
\end{table*}

\begin{table*}[]
    \centering
    \small
    \input{tables/appendixD/wmt_to100_tm1000_bleurt}
    \caption{Full WMT scenario with $|\mathcal{T}_{\Phi}| = 100$, and $|\mathcal{T}_M| = 1000$}
    \label{tab:wmt21_app4}
\end{table*}

\subsection{STB}
Tables~\ref{tab:stb_app1}, \ref{tab:stb_app2}, and \ref{tab:stb_app3} show the full evaluation of the three STB scenarios (see Section~\ref{sec:stb_sc}). Each table shows for each system the estimated $\alpha$ value in parentheses, and in each cell the $\epsilon$ value with $P(\alpha_1 > \alpha_2)$ in parentheses. All the pairs that are significantly distinguishable are put in bold. For the STB case, the six systems from the original paper are used: Blenderbot (BL)~\cite{roller-etal-2021-recipes}, Lost in Conversation (LiC)~\footnote{\url{https://github.com/atselousov/transformer_chatbot}}, KVMemNN (KV)~\cite{dinan2020second}, Huggingface (HF)~\footnote{\url{https://github.com/huggingface/transfer-learning-conv-ai}}, Bert-Rank (BR)~\cite{deriu-etal-2020-spot}, and Seq2Seq-NN (S2S)~\cite{deriu-etal-2020-spot}. Note that BR and S2S were custom trained baseline by the STB authors. In the STB case almost all pairs of systems are significantly distinguished, which is in line with the theory and the original STB paper. Our theory reveals that this is mostly due to the fact that the difference in $\alpha$ between the TGs is large and not many samples are needed for discriminating. 

\begin{table*}[]
    \centering
    \small
    \input{tables/appendixD/stb_to600_tm0}
    \caption{Full STB scenario with $|\mathcal{T}_{\Phi}| = 600$, and $|\mathcal{T}_M| = 0$}
    \label{tab:stb_app1}
\end{table*}

\begin{table*}[]
    \centering
    \small
    \input{tables/appendixD/stb_to100_tm0}
    \caption{Full STB scenario with $|\mathcal{T}_{\Phi}| = 100$, and $|\mathcal{T}_M| = 0$}
    \label{tab:stb_app2}
\end{table*}

\begin{table*}[]
    \centering
    \small
    \input{tables/appendixD/stb_to100_tm10000_usr_ret}
    \caption{Full STB scenario with $|\mathcal{T}_{\Phi}| = 100$, and $|\mathcal{T}_M| = 10000$}
    \label{tab:stb_app3}
\end{table*}

\section{Full Theory Tables}\label{app:tab_theory}
Tables~\ref{tbl:eps1},~\ref{tbl:eps2},~\ref{tbl:eps3}, and~\ref{tbl:eps4} show the distinguishable $\epsilon$ values for different combinations of $|\mathcal{T}_{\Phi}|$ and $|\mathcal{T}_M|$. Each table has different combinations of $\rho$, and $\eta$ values. For each table, we assume that $|\mathcal{T}_{\Phi}| = |\mathcal{T}_{\rho, \eta}|$. These tables can be used as guidelines for deciding on the number of human and automated ratings needed for different automated metric performances. 
 
\begin{table*}[ht!]
    \centering
    \small
    \input{tables/alpha60_rho70_eta70}
    \caption{Estimated $\epsilon_{\gamma}$ for $\alpha = 0.60$, $\rho = 0.70$, $\eta = 0.70$, and $\gamma = 0.05$}
    \label{tbl:eps1}
\end{table*}

\begin{table*}[ht!]
    \centering
    \small
    \input{tables/alpha60_rho90_eta90}
    \caption{Estimated $\epsilon_{\gamma}$ for $\alpha = 0.60$, $\rho = 0.90$, $\eta = 0.90$, and $\gamma = 0.05$}
    \label{tbl:eps2}
\end{table*}

\begin{table*}[ht!]
    \centering
    \small
    \input{tables/alpha60_rho99_eta99}
    \caption{Estimated $\epsilon_{\gamma}$ for $\alpha = 0.60$, $\rho = 0.99$, $\eta = 0.99$, and $\gamma = 0.05$}
    \label{tbl:eps3}
\end{table*}

\begin{table*}[ht!]
    \centering
    \small
    \input{tables/alpha60_rho51_eta51}
    \caption{Estimated $\epsilon_{\gamma}$ for $\alpha = 0.60$, $\rho = 0.51$, $\eta = 0.51$, and $\gamma = 0.05$}
    \label{tbl:eps4}
\end{table*}

\end{document}

%% file: tables/alpha60_rho70_eta70_fixed.tex
\begin{tabular}{ c | c | c  c c c c }
    \multicolumn{ 7 }{c}{$\#$Automated Ratings} \\ \hline
    \multirow{ 9 }{*}{\STAB{\rotatebox[origin=c]{90}{$\#$Human Ratings}}}
    &  & 0  & $1k$  &  $5k$  & $10k$  & $50k$    \\ \hline
    
        & 0  & 1.000  & 0.109  &  0.049  & 0.035  & 0.015    \\
    
        & 10  & 0.379  & 0.106  &  0.049  & 0.034  & 0.015    \\
    
    
        & 100  & 0.134  & 0.085  &  0.046  & 0.033  & 0.015    \\
    
    
    
        & $1k$  & 0.043  & 0.040  & 0.032  & 0.027  & 0.015    \\
    
        & $2.5k$  & 0.027  & 0.026    & 0.024  & \textbf{0.020}  & 0.013   \\
    
        & $5k$  & \textbf{0.019}  & 0.019  &  0.018  & 0.017  & 0.012  \\
    
\end{tabular}

%% file: tables/showcase1/bleurt_full_test.tex
\begin{tabular}{ l | c c  }
\multicolumn{ 3 }{c}{$|\mathcal{T}_{\Phi}| = 527, |\mathcal{T_M}| = 0$} \\ \hline
$\pi(\alpha_1) - \pi(\alpha_2)$   & $\epsilon$   & $P(\alpha_1 > \alpha_2)$   \\ \hline
FBAI(0.67) - VT (0.64)  & 0.02  & 0.798   \\
FBAI(0.67) - HU (0.58)  & \textbf{0.09}  & 0.998  \\
VT(0.64) - HU (0.58)    & \textbf{0.06}  & 0.978  \\\hline
\multicolumn{ 3 }{c}{$|\mathcal{T}_{\Phi}| = 100, |\mathcal{T_M}| = 0$} \\ \hline
$\pi(\alpha_1) - \pi(\alpha_2)$   & $\epsilon$   & $P(\alpha_1 > \alpha_2)$   \\ \hline
FBAI(0.67) - VT (0.65)  & 0.02  & 0.615   \\
FBAI(0.67) - HU (0.58)  & 0.09  & 0.904  \\
VT(0.65) - HU (0.58)    & 0.07  & 0.843  \\\hline
\multicolumn{ 3 }{c}{SacreBLEU: $|\mathcal{T}_{\Phi}| = 100, |\mathcal{T_M}| = 1000$} \\ \hline
$\pi(\alpha_1) - \pi(\alpha_2)$   & $\epsilon$   & $P(\alpha_1 > \alpha_2)$   \\ \hline
FBAI(0.67) - VT (0.64)  & 0.02  & 0.631   \\
FBAI(0.67) - HU (0.57)  & 0.09  & 0.918  \\
VT(0.64) - HU (0.57)    & 0.07  & 0.854  \\\hline
\multicolumn{ 3 }{c}{BleuRT: $|\mathcal{T}_{\Phi}| = 100, |\mathcal{T_M}| = 1000$} \\ \hline
$\pi(\alpha_1) - \pi(\alpha_2)$   & $\epsilon$   & $P(\alpha_1 > \alpha_2)$   \\ \hline
FBAI(0.66) - VT (0.62)  & 0.04  & 0.742   \\
FBAI(0.66) - HU (0.56)  & 0.09  & 0.933  \\
VT(0.62) - HU (0.56)    & 0.05  & 0.801  \\\hline
\end{tabular}

%% file: tables/showcase2/full_table.tex
\begin{tabular}{ l | c c  }
\multicolumn{ 3 }{c}{$|\mathcal{T}_{\Phi}| = 600, |\mathcal{T_M}| = 0$} \\ \hline
$\pi(\alpha_1) - \pi(\alpha_2)$   & $\epsilon$   & $P(\alpha_1 > \alpha_2)$   \\ \hline
BL (0.38) - LiC  (0.30)  & 0.08  & 0.999   \\
BL (0.30) - KV (0.24)  & 0.13  & 1.000  \\
LiC (0.30) - KV (0.24)    & 0.06  & 0.989  \\\hline
\multicolumn{ 3 }{c}{$|\mathcal{T}_{\Phi}| = 100, |\mathcal{T_M}| = 0$} \\ \hline
$\pi(\alpha_1) - \pi(\alpha_2)$   & $\epsilon$   & $P(\alpha_1 > \alpha_2)$   \\ \hline
BL (0.38) - LiC  (0.30)  & 0.08  & 0.882   \\
BL (0.38) - KV (0.25)  & 0.14  & 0.983  \\
LiC (0.30) - KV (0.25)    & 0.06  & 0.827  \\\hline
\multicolumn{ 3 }{c}{USR: $|\mathcal{T}_{\Phi}| = 100, |\mathcal{T_M}| = 10000$} \\ \hline
$\pi(\alpha_1) - \pi(\alpha_2)$   & $\epsilon$   & $P(\alpha_1 > \alpha_2)$   \\ \hline
BL (0.36) - LiC  (0.28)  & 0.08  & 0.889   \\
BL (0.36) - KV (0.22)  & 0.13  & 0.989  \\
LiC (0.28) - KV (0.22)    & 0.06  & 0.851  \\\hline
\end{tabular}

%% file: tables/appendixD/wmt_to527_tm0.tex
\begin{tabular}{ c | c c c c c c c }
     & FBAI ( 0.67 )   & VT-G ( 0.64 )   & OW ( 0.64 )   & NE ( 0.64 )   & VT-A ( 0.61 )   & UE ( 0.59 )   & HU ( 0.58 )   \\ \hline
    
        FBAI ( 0.67 )  & -  & 0.02 (0.798)  & 0.03 (0.848)  & 0.03 (0.862)  & 0.05 (0.968)  & \textbf{0.08 (0.997)}  & \textbf{0.09 (0.998)}  \\
    
        VT-G ( 0.64 )  & -0.02 (0.197)  & -  & 0.01 (0.573)  & 0.01 (0.598)  & 0.03 (0.844)  & 0.06 (0.971)  & \textbf{0.06 (0.978)}  \\
    
        OW ( 0.64 )  & -0.03 (0.148)  & -0.01 (0.420)  & -  & 0.00 (0.522)  & 0.02 (0.794)  & 0.05 (0.955)  & 0.05 (0.966)  \\
    
        NE ( 0.64 )  & -0.03 (0.134)  & -0.01 (0.395)  & -0.00 (0.471)  & -  & 0.02 (0.775)  & 0.05 (0.949)  & 0.05 (0.961)  \\
    
        VT-A ( 0.61 )  & -0.05 (0.031)  & -0.03 (0.152)  & -0.02 (0.202)  & -0.02 (0.220)  & -  & 0.03 (0.808)  & 0.03 (0.840)  \\
    
        UE ( 0.59 )  & \textbf{-0.08 (0.003)}  & -0.06 (0.028)  & -0.05 (0.043)  & -0.05 (0.049)  & -0.03 (0.187)  & -  & 0.00 (0.546)  \\
    
        HU ( 0.58 )  & \textbf{-0.09 (0.002)}  & \textbf{-0.06 (0.021)}  & -0.05 (0.033)  & -0.05 (0.038)  & -0.03 (0.156)  & -0.00 (0.447)  & -  \\
    
\end{tabular}

%% file: tables/appendixD/wmt_to100_tm0.tex
\begin{tabular}{ c | c c c c c c c }
     & FBAI ( 0.67 )   & VT-G ( 0.65 )   & NE ( 0.64 )   & OW ( 0.64 )   & VT-A ( 0.61 )   & UE ( 0.59 )   & HU ( 0.58 )   \\ \hline
    
        FBAI ( 0.67 )  & -  & 0.02 (0.615)  & 0.03 (0.670)  & 0.03 (0.670)  & 0.06 (0.809)  & 0.08 (0.877)  & 0.09 (0.904)  \\
    
        VT-G ( 0.65 )  & -0.02 (0.382)  & -  & 0.01 (0.557)  & 0.01 (0.557)  & 0.04 (0.719)  & 0.06 (0.807)  & 0.07 (0.843)  \\
    
        NE ( 0.64 )  & -0.03 (0.327)  & -0.01 (0.440)  & -  & 0.00 (0.499)  & 0.03 (0.667)  & 0.05 (0.764)  & 0.06 (0.805)  \\
    
        OW ( 0.64 )  & -0.03 (0.327)  & -0.01 (0.440)  & 0.00 (0.499)  & -  & 0.03 (0.667)  & 0.05 (0.764)  & 0.06 (0.805)  \\
    
        VT-A ( 0.61 )  & -0.06 (0.189)  & -0.04 (0.279)  & -0.03 (0.330)  & -0.03 (0.330)  & -  & 0.02 (0.612)  & 0.03 (0.665)  \\
    
        UEdin ( 0.59 )  & -0.08 (0.121)  & -0.06 (0.191)  & -0.05 (0.234)  & -0.05 (0.234)  & -0.02 (0.386)  & -  & 0.01 (0.555)  \\
    
        HU ( 0.58 )  & -0.09 (0.095)  & -0.07 (0.155)  & -0.06 (0.193)  & -0.06 (0.193)  & -0.03 (0.332)  & -0.01 (0.442)  & -  \\
    
\end{tabular}

%% file: tables/appendixD/wmt_to100_tm1000_sentbleu.tex
\begin{tabular}{ c | c c c c c c c }
     & FBAI ( 0.67 )   & VT-G ( 0.64 )   & NE ( 0.63 )   & OW ( 0.63 )   & VT-A ( 0.61 )   & UE ( 0.58 )   & HU ( 0.57 )   \\ \hline
    
        FBAI ( 0.67 )  & -  & 0.02 (0.631)  & 0.03 (0.689)  & 0.04 (0.713)  & 0.06 (0.817)  & 0.08 (0.896)  & 0.09 (0.918)  \\
    
        VT-G ( 0.64 )  & -0.02 (0.366)  & -  & 0.01 (0.560)  & 0.01 (0.588)  & 0.04 (0.713)  & 0.06 (0.821)  & 0.07 (0.854)  \\
    
        NE ( 0.63 )  & -0.03 (0.309)  & -0.01 (0.437)  & -  & 0.00 (0.527)  & 0.03 (0.658)  & 0.05 (0.779)  & 0.06 (0.817)  \\
    
        OW ( 0.63 )  & -0.04 (0.284)  & -0.01 (0.409)  & -0.00 (0.470)  & -  & 0.02 (0.631)  & 0.05 (0.757)  & 0.06 (0.798)  \\
    
        VT-A ( 0.61 )  & -0.06 (0.181)  & -0.04 (0.284)  & -0.03 (0.339)  & -0.02 (0.366)  & -  & 0.02 (0.643)  & 0.03 (0.694)  \\
    
        UE ( 0.58 )  & -0.08 (0.103)  & -0.06 (0.177)  & -0.05 (0.219)  & -0.05 (0.240)  & -0.02 (0.354)  & -  & 0.01 (0.555)  \\
    
        HU ( 0.57 )  & -0.09 (0.081)  & -0.07 (0.144)  & -0.06 (0.181)  & -0.06 (0.200)  & -0.03 (0.303)  & -0.01 (0.442)  & -  \\
    
\end{tabular}

%% file: tables/appendixD/wmt_to100_tm1000_bleurt.tex
\begin{tabular}{ c | c c c c c c c }
     & FBAI ( 0.66 )   & VT-G ( 0.62 )   & NE ( 0.61 )   & OW ( 0.61 )   & VT-A ( 0.58 )   & UE ( 0.57 )   & HU ( 0.56 )   \\ \hline
    
        FBAI ( 0.66 )  & -  & 0.04 (0.742)  & 0.05 (0.787)  & 0.05 (0.779)  & 0.07 (0.888)  & 0.09 (0.916)  & 0.09 (0.933)  \\
    
        VT-G ( 0.62 )  & -0.04 (0.256)  & -  & 0.01 (0.538)  & 0.01 (0.548)  & 0.03 (0.705)  & 0.05 (0.768)  & 0.05 (0.801)  \\
    
        NE ( 0.61 )  & -0.05 (0.211)  & -0.01 (0.458)  & -  & 0.00 (0.512)  & 0.03 (0.684)  & 0.04 (0.753)  & 0.05 (0.790)  \\
    
        OW ( 0.61 )  & -0.05 (0.219)  & -0.01 (0.449)  & -0.00 (0.485)  & -  & 0.02 (0.658)  & 0.04 (0.727)  & 0.04 (0.762)  \\
    
        VT-A ( 0.58 )  & -0.07 (0.111)  & -0.03 (0.292)  & -0.03 (0.313)  & -0.02 (0.339)  & -  & 0.01 (0.587)  & 0.02 (0.629)  \\
    
        UE ( 0.57 )  & -0.09 (0.083)  & -0.05 (0.230)  & -0.04 (0.244)  & -0.04 (0.271)  & -0.01 (0.410)  & -  & 0.01 (0.539)  \\
    
        HU ( 0.56 )  & -0.09 (0.066)  & -0.05 (0.197)  & -0.05 (0.208)  & -0.04 (0.236)  & -0.02 (0.368)  & -0.01 (0.458)  & -  \\
    
\end{tabular}

%% file: tables/appendixD/stb_to600_tm0.tex
\begin{tabular}{ c | c c c c c c }
     & BL ( 0.38 )   & LiC ( 0.30 )   & KV ( 0.24 )   & HF ( 0.18 )   & BR ( 0.07 )   & S2S ( 0.04 )   \\ \hline
    
        BL ( 0.38 )  & -  & \textbf{0.08 (0.999)}  & \textbf{0.13 (1.000)}  & \textbf{0.20 (1.000)}  & \textbf{0.31 (1.000)}  & \textbf{0.34 (1.000)}  \\
    
        LiC ( 0.30 )  & \textbf{-0.08 (0.001)}  & -  & \textbf{0.06 (0.989)}  & \textbf{0.12 (1.000)}  & \textbf{0.23 (1.000)}  & \textbf{0.26 (1.000)}  \\
    
        KV ( 0.24 )  & \textbf{-0.13 (0.000)}  & \textbf{-0.06 (0.010)}  & -  & \textbf{0.07 (0.998)}  & \textbf{0.18 (1.000)}  & \textbf{0.20 (1.000)}  \\
    
        HF ( 0.18 )  & \textbf{-0.20 (0.000)} & \textbf{-0.12 (0.000)}  & \textbf{-0.07 (0.002)}  & -  & \textbf{0.11 (1.000)}  & \textbf{0.14 (1.000)}  \\
    
        BR ( 0.07 )  & \textbf{-0.31 (0.000)} & \textbf{-0.23 (0.000)}  & \textbf{-0.18 (0.000)}  & \textbf{-0.11 (0.000)}  & -  & 0.02 (0.974)  \\
    
        S2S ( 0.04 )  & \textbf{-0.34 (0.000)} & \textbf{-0.26 (0.000)}  & \textbf{-0.20 (0.000)}  & \textbf{-0.14 (0.000)}  & -0.02 (0.024)  & -  \\
    
\end{tabular}

%% file: tables/appendixD/stb_to100_tm0.tex
\begin{tabular}{ c | c c c c c c }
     & BL ( 0.38 )   & LiC ( 0.30 )   & KV ( 0.25 )   & HF ( 0.19 )   & BR ( 0.07 )   & S2S ( 0.05 )   \\ \hline
    
        BL ( 0.38 )  & -  & 0.08 (0.882)  & \textbf{0.14 (0.983)}  & \textbf{0.20 (0.999)}  & \textbf{0.31 (1.000)}  & \textbf{0.33 (1.000)}  \\
    
        LiC ( 0.30 )  & -0.08 (0.117)  & -  & 0.06 (0.827)  & \textbf{0.12 (0.976)}  & \textbf{0.24 (1.000)}  & \textbf{0.25 (1.000)}  \\
    
        KV ( 0.25 )  & \textbf{-0.14 (0.016)}  & -0.06 (0.170)  & -  & 0.06 (0.848)  & \textbf{0.18 (1.000)}  & \textbf{0.20 (1.000)}  \\
    
        HF ( 0.19 )  & \textbf{-0.20 (0.001)}  & \textbf{-0.12 (0.024)}  & -0.06 (0.150)  & -  & \textbf{0.12 (0.995)}  & \textbf{0.14 (0.999)}  \\
    
        BR ( 0.07 )  & \textbf{-0.31 (0.000)}  & \textbf{-0.24 (0.000)}  & \textbf{-0.18 (0.000)}  & \textbf{-0.12 (0.005)}  & -  & 0.02 (0.729)  \\
    
        S2S ( 0.05 )  & \textbf{-0.33 (0.000)}  &\textbf{ -0.25 (0.000)}  & \textbf{-0.20 (0.000)}  & \textbf{-0.14 (0.001)} & -0.02 (0.266)  & -  \\
    
\end{tabular}

%% file: tables/appendixD/stb_to100_tm10000_usr_ret.tex
\begin{tabular}{ c | c c c c c c }
     & BL ( 0.36 )   & LiC ( 0.28 )   & KV ( 0.22 )   & HF ( 0.15 )   & BR ( 0.06 )   & S2S ( 0.05 )   \\ \hline
    
        BL ( 0.36 )  & -  & 0.08 (0.889)  & \textbf{0.13 (0.989)}  & \textbf{0.21 (1.000)}  & \textbf{0.30 (1.000)}  & \textbf{0.31 (1.000)}  \\
    
        LiC ( 0.28 )  & -0.08 (0.109)  & -  & 0.06 (0.851)  & \textbf{0.13 (0.994)}  & \textbf{0.22 (1.000)}  & \textbf{0.23 (1.000)}  \\
    
        KV ( 0.22 )  & \textbf{-0.13 (0.010)}  & -0.06 (0.147)  & -  & 0.07 (0.935)  & \textbf{0.16 (1.000)}  & \textbf{0.17 (1.000)}  \\
    
        HF ( 0.15 )  & \textbf{-0.21 (0.000)}  & \textbf{-0.13 (0.006)} & -0.07 (0.064)  & -  & \textbf{0.09 (0.994)}  & \textbf{0.10 (0.997)}  \\
    
        BR ( 0.06 )  & \textbf{-0.30 (0.000)}  & \textbf{-0.22 (0.000)}  & \textbf{-0.16 (0.000)}  & \textbf{-0.09 (0.006)}  & -  & 0.01 (0.628)  \\
    
        S2S ( 0.05 )  & \textbf{-0.31 (0.000)}  & \textbf{-0.23 (0.000)}  & \textbf{-0.17 (0.000)}  & \textbf{-0.10 (0.003)}  & -0.01 (0.365)  & -  \\
    
\end{tabular}

%% file: tables/alpha60_rho70_eta70.tex
\begin{tabular}{ c | c | c c c c c c c }
    \multicolumn{ 9 }{c}{$|\mathcal{T_M}|$} \\ \hline
    \multirow{ 9 }{*}{\STAB{\rotatebox[origin=c]{90}{$|\mathcal{T_{\rho, \eta}}| = |\mathcal{T_O}|$}}}
    &  & 0  & 1000  & 2500  & 5000  & 10000  & 50000  & 100000  \\ \hline
    
        & 0  & 1.000  & 0.734  & 0.733  & 0.733  & 0.733  & 0.733  & 0.733  \\
    
        & 100  & 0.134  & 0.124  & 0.124  & 0.123  & 0.123  & 0.123  & 0.123  \\
    
        & 250  & 0.085  & 0.080  & 0.079  & 0.079  & 0.079  & 0.079  & 0.079  \\
    
        & 500  & 0.061  & 0.057  & 0.057  & 0.056  & 0.056  & 0.056  & 0.056  \\
    
        & 1000  & 0.043  & 0.041  & 0.040  & 0.040  & 0.040  & 0.040  & 0.039  \\
    
        & 2500  & 0.027  & 0.027  & 0.026  & 0.026  & 0.025  & 0.025  & 0.025  \\
    
        & 5000  & 0.019  & 0.019  & 0.019  & 0.018  & 0.018  & 0.018  & 0.018  \\
    
        & 10000  & 0.014  & 0.013  & 0.013  & 0.013  & 0.013  & 0.013  & 0.013  \\
    
\end{tabular}

%% file: tables/alpha60_rho90_eta90.tex
\begin{tabular}{ c | c | c c c c c c c }
    \multicolumn{ 9 }{c}{$|\mathcal{T_M}|$} \\ \hline
    \multirow{ 9 }{*}{\STAB{\rotatebox[origin=c]{90}{$|\mathcal{T_{\rho, \eta}}| = |\mathcal{T_O}|$}}}
    &  & 0  & 1000  & 2500  & 5000  & 10000  & 50000  & 100000  \\ \hline
    
        & 0  & 1.000  & 0.739  & 0.738  & 0.738  & 0.738  & 0.738  & 0.738  \\
    
        & 100  & 0.134  & 0.091  & 0.088  & 0.087  & 0.086  & 0.086  & 0.086  \\
    
        & 250  & 0.085  & 0.061  & 0.057  & 0.055  & 0.054  & 0.053  & 0.053  \\
    
        & 500  & 0.061  & 0.046  & 0.042  & 0.040  & 0.039  & 0.038  & 0.037  \\
    
        & 1000  & 0.043  & 0.036  & 0.032  & 0.030  & 0.028  & 0.027  & 0.026  \\
    
        & 2500  & 0.027  & 0.025  & 0.022  & 0.021  & 0.019  & 0.017  & 0.017  \\
    
        & 5000  & 0.019  & 0.018  & 0.017  & 0.016  & 0.015  & 0.013  & 0.012  \\
    
        & 10000  & 0.014  & 0.013  & 0.013  & 0.012  & 0.011  & 0.009  & 0.009  \\
    
\end{tabular}

%% file: tables/alpha60_rho99_eta99.tex
\begin{tabular}{ c | c | c c c c c c c }
    \multicolumn{ 9 }{c}{$|\mathcal{T_M}|$} \\ \hline
    \multirow{ 9 }{*}{\STAB{\rotatebox[origin=c]{90}{$|\mathcal{T_{\rho, \eta}}| = |\mathcal{T_O}|$}}}
    &  & 0  & 1000  & 2500  & 5000  & 10000  & 50000  & 100000  \\ \hline
    
        & 0  & 1.000  & 0.742  & 0.742  & 0.742  & 0.742  & 0.742  & 0.742  \\
    
        & 100  & 0.134  & 0.059  & 0.051  & 0.048  & 0.046  & 0.045  & 0.045  \\
    
        & 250  & 0.085  & 0.044  & 0.035  & 0.030  & 0.027  & 0.025  & 0.024  \\
    
        & 500  & 0.061  & 0.037  & 0.028  & 0.023  & 0.019  & 0.016  & 0.015  \\
    
        & 1000  & 0.043  & 0.031  & 0.024  & 0.020  & 0.016  & 0.011  & 0.010  \\
    
        & 2500  & 0.027  & 0.023  & 0.020  & 0.016  & 0.013  & 0.008  & 0.007  \\
    
        & 5000  & 0.019  & 0.018  & 0.016  & 0.014  & 0.012  & 0.007  & 0.006  \\
    
        & 10000  & 0.014  & 0.013  & 0.012  & 0.011  & 0.010  & 0.006  & 0.005  \\
    
\end{tabular}

%% file: tables/alpha60_rho51_eta51.tex
\begin{tabular}{ c | c | c c c c c c c }
    \multicolumn{ 9 }{c}{$|\mathcal{T_M}|$} \\ \hline
    \multirow{ 9 }{*}{\STAB{\rotatebox[origin=c]{90}{$|\mathcal{T_{\rho, \eta}}| = |\mathcal{T_O}|$}}}
    &  & 0  & 1000  & 2500  & 5000  & 10000  & 50000  & 100000  \\ \hline
    
        & 0  & 1.000  & 0.732  & 0.732  & 0.732  & 0.732  & 0.732  & 0.732  \\
    
        & 100  & 0.134  & 0.133  & 0.133  & 0.133  & 0.133  & 0.133  & 0.133  \\
    
        & 250  & 0.085  & 0.085  & 0.085  & 0.085  & 0.085  & 0.085  & 0.085  \\
    
        & 500  & 0.061  & 0.061  & 0.060  & 0.060  & 0.060  & 0.060  & 0.060  \\
    
        & 1000  & 0.043  & 0.043  & 0.043  & 0.043  & 0.043  & 0.043  & 0.043  \\
    
        & 2500  & 0.027  & 0.027  & 0.027  & 0.027  & 0.027  & 0.027  & 0.027  \\
    
        & 5000  & 0.019  & 0.019  & 0.019  & 0.019  & 0.019  & 0.019  & 0.019  \\
    
        & 10000  & 0.014  & 0.014  & 0.014  & 0.014  & 0.014  & 0.014  & 0.014  \\
    
\end{tabular}